\documentclass{configuration/antgroup}

\PassOptionsToPackage{numbers, compress}{natbib}
\usepackage{configuration/antgroup}

\usepackage{hyperref}
\usepackage{url}
\usepackage{enumitem}
\usepackage[most]{tcolorbox}
\usepackage{listings}
\usepackage{fontawesome5}

\usepackage{amsmath,amssymb,amsthm}

\newenvironment{talign*}
{\csname align*\endcsname}
{\endalign}

\usepackage[utf8]{inputenc}         %
\usepackage[T1]{fontenc}            %
\usepackage{url}                    %
\usepackage{booktabs}               %
\usepackage{amsfonts}               %
\usepackage{nicefrac}               %
\usepackage{microtype}              %
\usepackage{xcolor}                 %
\usepackage{algorithm}
\usepackage{algorithmic}
\usepackage{graphicx}
\usepackage{subcaption}
\usepackage[flushleft]{threeparttable}
\usepackage{float}
\usepackage{multirow}
\usepackage{xspace}
\usepackage{enumitem}
\usepackage[font=small]{caption}
\usepackage{autobreak}
\usepackage{sidecap}
\usepackage{wrapfig}
\usepackage{bbding}
\usepackage[toc, page, header]{appendix}
\usepackage{tikz}
\usepackage{xcolor}
\usepackage{pifont}
\usepackage{mdframed}
\usepackage{colortbl}
\usepackage{arydshln}

\hypersetup{
	colorlinks=true,
	linkcolor=blue,
	citecolor=blue,
	urlcolor=blue
}

\definecolor{coral}{RGB}{255,127,80}
\definecolor{darkgreen}{RGB}{0,100,0}
\definecolor{darkyellow}{RGB}{204,153,0}
\definecolor{salmon}{RGB}{250,128,114}
\definecolor{darkred}{RGB}{150,0,0}
\newcommand{\darkredtext}[1]{{\color{darkred}#1}}

\newcommand{\secref}[1]{\hyperref[#1]{\darkredtext{Sec.~\ref*{#1}}}}
\newcommand{\thmref}[1]{\hyperref[#1]{\darkredtext{Thm.~\ref*{#1}}}}
\newcommand{\defref}[1]{\hyperref[#1]{\darkredtext{Def.~\ref*{#1}}}}
\newcommand{\propref}[1]{\hyperref[#1]{\darkredtext{Prop.~\ref*{#1}}}}
\newcommand{\assumpref}[1]{\hyperref[#1]{\darkredtext{Assump.~\ref*{#1}}}}
\newcommand{\remarkref}[1]{\hyperref[#1]{\darkredtext{Rem.~\ref*{#1}}}}
\newcommand{\hypref}[1]{\hyperref[#1]{\darkredtext{Hyp.~\ref*{#1}}}}
\newcommand{\conjref}[1]{\hyperref[#1]{\darkredtext{Conj.~\ref*{#1}}}}
\newcommand{\lemref}[1]{\hyperref[#1]{\darkredtext{Lem.~\ref*{#1}}}}
\newcommand{\corref}[1]{\hyperref[#1]{\darkredtext{Cor.~\ref*{#1}}}}
\newcommand{\noteref}[1]{\hyperref[#1]{\darkredtext{Nota.~\ref*{#1}}}}
\newcommand{\claimref}[1]{\hyperref[#1]{\darkredtext{Clm.~\ref*{#1}}}}
\newcommand{\obsref}[1]{\hyperref[#1]{\darkredtext{Obs.~\ref*{#1}}}}
\newcommand{\algref}[1]{\hyperref[#1]{\darkredtext{Alg.~\ref*{#1}}}}
\newcommand{\figref}[1]{\hyperref[#1]{\darkredtext{Fig.~\ref*{#1}}}}
\newcommand{\tabref}[1]{\hyperref[#1]{\darkredtext{Tab.~\ref*{#1}}}}
\newcommand{\appref}[1]{\hyperref[#1]{\darkredtext{App.~\ref*{#1}}}}

\newtheoremstyle{custom}
{1pt} %
{1pt} %
{\itshape} %
{} %
{\bfseries} %
{} %
{ } %
{\thmname{#1} \thmnumber{#2} \thmnote{(#3)} . } %

\theoremstyle{custom}

\newtheorem{innerdefinition}{Definition}
\newtheorem{innerproposition}{Proposition}
\newtheorem{innerassumption}{Assumption}
\newtheorem{innerremark}{Remark}
\newtheorem{innertheorem}{Theorem}
\newtheorem{innerhypothesis}{Hypothesis}
\newtheorem{innerconjecture}{Conjecture}
\newtheorem{innerlemma}{Lemma}
\newtheorem{innercorollary}{Corollary}
\newtheorem{innernotation}{Notation}
\newtheorem{innerclaim}{Claim}
\newtheorem{innerproblem}{Problem}

\newtheorem{innerobservation}{Observation}

\newmdenv[
	backgroundcolor=gray!10,
	linecolor=gray!100,
	linewidth=0.8pt,
	skipabove=2pt,
	skipbelow=2pt,
	innertopmargin=10pt,
	innerbottommargin=5pt,
	innerleftmargin=5pt,
	innerrightmargin=5pt,
]{definitionframe}

\newmdenv[
	backgroundcolor=blue!10,
	linecolor=blue!100,
	linewidth=0.8pt,
	skipabove=2pt,
	skipbelow=2pt,
	innertopmargin=10pt,
	innerbottommargin=5pt,
	innerleftmargin=5pt,
	innerrightmargin=5pt,
]{propositionframe}

\newmdenv[
	backgroundcolor=green!10,
	linecolor=green!100,
	linewidth=0.8pt,
	skipabove=2pt,
	skipbelow=2pt,
	innertopmargin=10pt,
	innerbottommargin=5pt,
	innerleftmargin=5pt,
	innerrightmargin=5pt,
]{assumptionframe}

\newmdenv[
	backgroundcolor=yellow!10,
	linecolor=yellow!100,
	linewidth=0.8pt,
	skipabove=2pt,
	skipbelow=2pt,
	innertopmargin=10pt,
	innerbottommargin=5pt,
	innerleftmargin=5pt,
	innerrightmargin=5pt,
]{remarkframe}

\newmdenv[
	backgroundcolor=red!10,
	linecolor=red!100,
	linewidth=0.8pt,
	skipabove=2pt,
	skipbelow=2pt,
	innertopmargin=10pt,
	innerbottommargin=5pt,
	innerleftmargin=5pt,
	innerrightmargin=5pt,
]{theoremframe}

\newmdenv[
	backgroundcolor=purple!10,
	linecolor=purple!100,
	linewidth=0.8pt,
	skipabove=2pt,
	skipbelow=2pt,
	innertopmargin=10pt,
	innerbottommargin=5pt,
	innerleftmargin=5pt,
	innerrightmargin=5pt,
]{hypothesisframe}

\newmdenv[
	backgroundcolor=orange!10,
	linecolor=orange!100,
	linewidth=0.8pt,
	skipabove=2pt,
	skipbelow=2pt,
	innertopmargin=10pt,
	innerbottommargin=5pt,
	innerleftmargin=5pt,
	innerrightmargin=5pt,
]{conjectureframe}

\newmdenv[
	backgroundcolor=cyan!10,
	linecolor=cyan!100,
	linewidth=0.8pt,
	skipabove=2pt,
	skipbelow=2pt,
	innertopmargin=10pt,
	innerbottommargin=5pt,
	innerleftmargin=5pt,
	innerrightmargin=5pt,
]{lemmaframe}

\newmdenv[
	backgroundcolor=magenta!10,
	linecolor=magenta!100,
	linewidth=0.8pt,
	skipabove=2pt,
	skipbelow=2pt,
	innertopmargin=10pt,
	innerbottommargin=5pt,
	innerleftmargin=5pt,
	innerrightmargin=5pt,
]{corollaryframe}

\newmdenv[
	backgroundcolor=pink!10,
	linecolor=pink!100,
	linewidth=0.8pt,
	skipabove=2pt,
	skipbelow=2pt,
	innertopmargin=10pt,
	innerbottommargin=5pt,
	innerleftmargin=5pt,
	innerrightmargin=5pt,
]{notationframe}

\newmdenv[
	backgroundcolor=violet!10,
	linecolor=violet!100,
	linewidth=0.8pt,
	skipabove=2pt,
	skipbelow=2pt,
	innertopmargin=10pt,
	innerbottommargin=5pt,
	innerleftmargin=5pt,
	innerrightmargin=5pt,
]{claimframe}

\newmdenv[
	backgroundcolor=salmon!10,
	linecolor=salmon!100,
	linewidth=0.8pt,
	skipabove=2pt,
	skipbelow=2pt,
	innertopmargin=10pt,
	innerbottommargin=5pt,
	innerleftmargin=5pt,
	innerrightmargin=5pt,
]{problemframe}

\newmdenv[
	backgroundcolor=lavender!10,
	linecolor=lavender!100,
	linewidth=0.8pt,
	skipabove=2pt,
	skipbelow=2pt,
	innertopmargin=10pt,
	innerbottommargin=5pt,
	innerleftmargin=5pt,
	innerrightmargin=5pt,
]{observationframe}

\usepackage[most]{tcolorbox}

\newtcolorbox{modernbox}{
  enhanced,
  breakable,
  sharp corners,
  colback=gray!5,
  colframe=gray!80,
  leftrule=3pt,
  rightrule=0pt,
  toprule=0pt,
  bottomrule=0pt,
  before skip=10pt,
  after skip=10pt,
}

\usepackage[textwidth=3.5cm]{todonotes}

\definecolor{thinkcolor}{RGB}{138, 43, 226}
\definecolor{toolcolor}{RGB}{34, 139, 34}
\definecolor{toolresponsecolor}{RGB}{70, 130, 180}
\definecolor{answercolor}{RGB}{25, 25, 112}
\definecolor{usercolor}{RGB}{105, 105, 105}
\definecolor{groundtruthcolor}{RGB}{139, 69, 19}
\definecolor{querycolor}{RGB}{70, 130, 180}
\definecolor{tracecolor}{RGB}{34, 139, 34}

\newcommand{\ours}{\textsc{RODS}\xspace}
\newcommand{\improvement}[1]{\textcolor{red}{#1}}
\newcommand{\degradation}[1]{\textcolor{orange}{#1}}

\title{RODS: Reward-Driven Online Data Synthesis\\ for Multi-Turn Tool-Use Agents}
\renewcommand{\titletext}{Reward-Driven Online Data Synthesis for Multi-Turn Tool-Use Agents}

\author{
  Ruishan Fang$^{1,2,4,\ast}$\quad Siyuan Lu$^{1,2,3,4}$\quad Chenyi Zhuang$^{1,\dagger}$\quad Tao Lin$^{4,1,\dagger}$
}

\affiliation{$^1$Inclusion AI, Ant Group\quad}
\affiliation{$^2$Zhejiang University\quad}
\affiliation{$^3$Shanghai Innovation Institute\quad}
\affiliation{$^4$Westlake University\\}

\begin{document}

{\renewcommand\thefootnote{}
  \footnotetext{$^\ast$ This work was supported by Ant Group Research Intern Program.}
  \footnotetext{$^\dagger$Corresponding Authors.}
}

\maketitle

\begin{center}
  \vspace{-0.5em}
  \raisebox{-0.1em}{\includegraphics[height=0.9em]{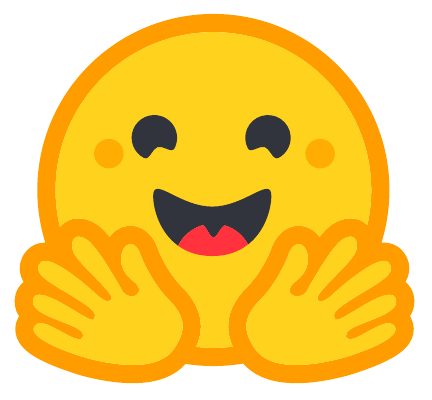}} \href{https://huggingface.co/RuishanFang/Qwen3-4B-RODS}{\texttt{Model}}\quad\quad\href{https://github.com/inclusionAI/AWorld-RL/tree/main/RODS}{\faGithub\ \texttt{AWorld-RL/RODS}}
  \vspace{1.5em}
\end{center}

\begin{abstract}

	Multi-turn tool-use RL is bottlenecked by the rapid depletion of informative samples in static datasets.
	We observe that the gradient signal in GRPO concentrates on tasks with the highest rollout reward variance, a consequence of the Popoviciu upper bound. Consequently, samples near the agent's \emph{capability boundary}---where successes and failures are roughly balanced---contribute disproportionately large policy gradients.
	As training progresses, this boundary continuously shifts, which gradually depletes the pool of informative samples in a static dataset.
	We propose \ours (\textbf{R}eward-driven \textbf{O}nline \textbf{D}ata \textbf{S}ynthesis) to resolve this depletion. \ours closes the loop between RL training and data generation by repurposing the progress reward variance as a practical, zero-cost boundary detector that requires no extra inference beyond the rollouts already computed for training.
	It continuously identifies such boundary samples, synthesizes new multi-turn variants matching their structural complexity (e.g., API topology and dependency depth) via a skill-aligned resampling pipeline, and manages a dynamic replay buffer that co-evolves with the policy.
	Starting from 400 human seeds and maintaining an active training pool of $\sim$800 samples, \ours achieves comparable performance to a 17K-sample offline pipeline while requiring roughly $20\times$ fewer trajectories, and improves over fixed-data RL and environment augmentation in our controlled setting.
	\looseness=-1
\end{abstract}

\begin{figure}[htbp]
	\centering
	\includegraphics[width=0.95\textwidth]{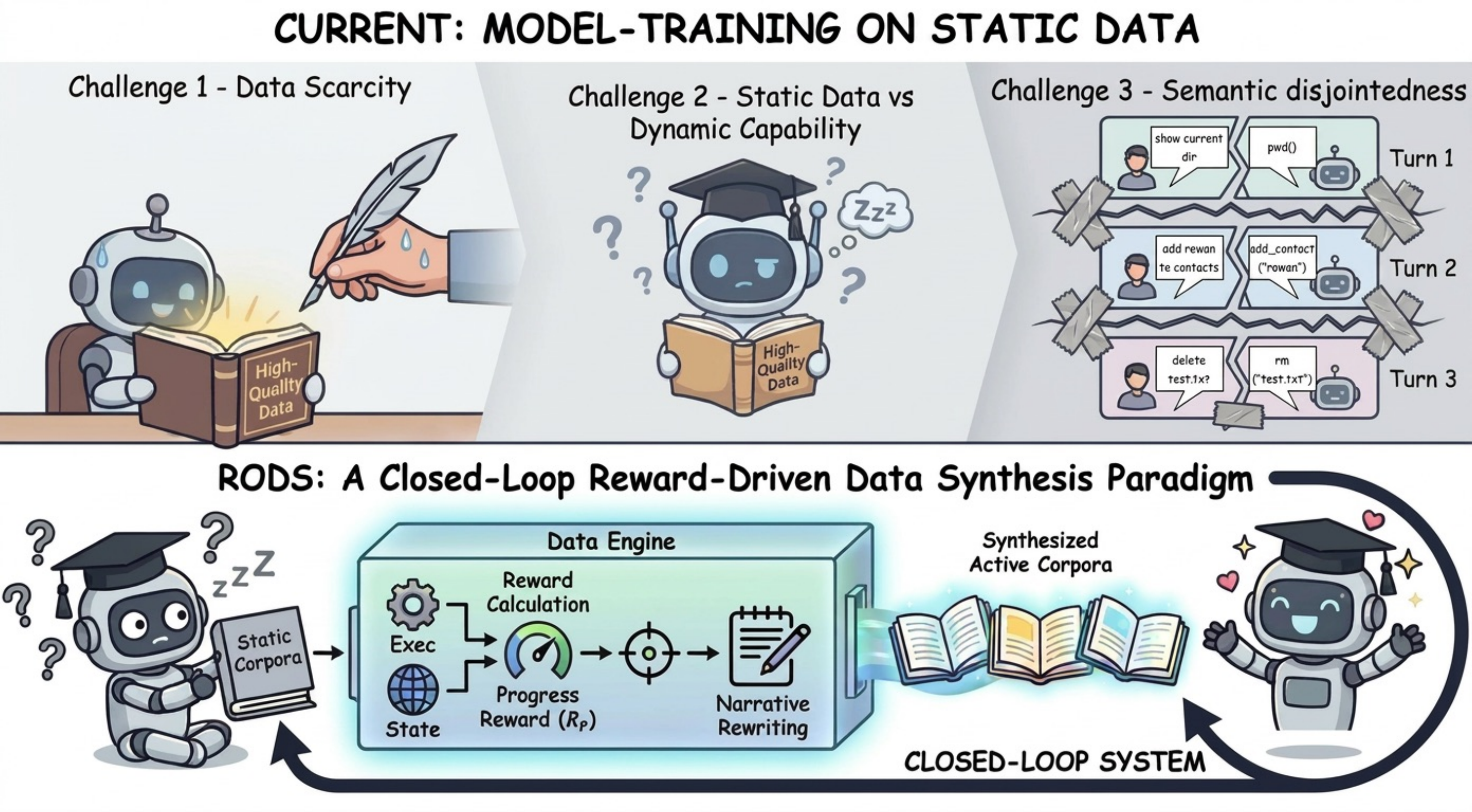}
	\vspace{-1em}
	\caption{\small
		\textbf{Current limitations in static data training vs.\ \ours dynamic synthesis paradigm.}
		\textbf{(a) Data scarcity ($\mathbb{C}1$):} High-quality multi-turn tool-use datasets require massive human annotation effort.
		\textbf{(b) Capability vs.\ static Data ($\mathbb{C}2$):} As the model learns, static data becomes mastered, leading to a loss of gradient signal.
		\textbf{(c) Semantic disjointedness ($\mathbb{C}3$):} Naively stitching single-turn queries creates disjointed interactions. \\
		\textbf{(d) The \ours solution:} A closed-loop data engine continuously ingests static data to detect the evolving capability boundary, synthesizing coherent, targeted active corpora where the model needs them.
	}
	\vspace{-1em}
	\label{fig:challenges}
\end{figure}

\section{Introduction}
Large Language Model (LLM)-based agents~\citep{wang2024survey,weng2023agent} have demonstrated strong potential in solving complex tasks by using external tools and environment interactions~\citep{anthropic2026claude,openai2026gpt54thinking}.
Recent work frames tool-use agent training as an RL problem (Agentic RL), optimizing policies through direct environment interaction~\citep{jin2025search,feng2025retool,wang2025ragen,liu2025exploratory}.
However, extending Agentic RL to multi-turn tool-use environments introduces unique challenges that have not been fully addressed by existing methods, as it simultaneously demands large training corpora and long-horizon structural coherence~\citep{patil2025bfcl,tbench,tau2bench}.

Current RL systems face three main challenges in this regime (see Figure~\ref{fig:challenges}).
First, high-quality multi-turn datasets are scarce ($\mathbb{C}1$) due to prohibitive annotation and validation costs, as seen in BFCL V3~\citep{patil2025bfcl} which contains only 800 samples.
Second, as agent capabilities evolve, static datasets suffer from shifting boundaries ($\mathbb{C}2$), leading to signal depletion and wasted compute on mastered or unreachable tasks~\citep{li2025adacurl,dai2025harder}.
Third, on-the-fly synthesis often causes semantic disjointedness ($\mathbb{C}3$) by lacking a unifying goal, which produces trajectories without coreference or coherence that fail to teach reliable reasoning.

Despite these hurdles, existing efforts generally fall into two categories: large-scale offline synthesis and online environment augmentation.
Offline pipelines~\citep{prabhakar2025apigen,toucan,xu2025funreasonmt} address data scarcity ($\mathbb{C}1$) by generating massive corpora upfront, yet they remain decoupled from the training loop and thus fail to track the model's evolving capability boundary ($\mathbb{C}2$). Conversely, RL training methods like EnvTuning~\citep{tuneenv} enable learning from minimal data but are ultimately constrained by the signal depletion of their fixed seed corpora ($\mathbb{C}2$). While online self-play and self-evolution approaches~\citep{toolr0,li2025closeloop,zhai2025agentevolver} theoretically close this loop, unconstrained zero-data generation frequently fails to maintain semantic coherence in complex multi-turn settings ($\mathbb{C}3$), leading to disjointed trajectories that offer little pedagogical value.

\begin{tcolorbox}[
		colback=salmon!10,
		colframe=salmon!80!black,
		arc=2mm,
		coltitle=white,
		left=1mm,
		right=1mm,
		top=0.8mm,
		bottom=0.8mm,
	]
	\centering
	\emph{How can we train multi-turn tool-use agents under extreme \textbf{data scarcity}, \\ by dynamically synthesizing data that strictly tracks the evolving \textbf{capability boundary} \\ while maintaining \textbf{multi-turn semantic coherence}?}
\end{tcolorbox}

We propose \ours (\textbf{R}eward-driven \textbf{O}nline \textbf{D}ata \textbf{S}ynthesis) to bridge this gap, a framework that tightly couples data generation with the RL training loop. Our approach rests on a simple insight: the Progress Reward in policy gradient methods like GRPO serves as a boundary detector that reuses existing rollout statistics---since rollouts are already computed for advantage estimation---because rollout variance is highest near the agent's capability boundary ($\mu \approx 0.5$, as suggested by the Popoviciu upper bound). By synthesizing data in this high-variance region and managing its lifecycle through a co-evolving replay buffer, \ours maintains a continuous stream of gradient-informative samples.
This paper makes three main contributions.
To address signal starvation inherent to static datasets ($\mathbb{C}2$), we propose \ours, a reward-driven boundary expansion method that repurposes the RL progress reward to identify and expand boundary tasks in real time.
To preserve multi-turn semantic coherence ($\mathbb{C}3$), we introduce a skill-aligned resampling synthesis pipeline that anchors novel trajectories to the complexity profiles of verified seeds.
Instead of simple entity substitution, it preserves the functional dependency structure of the seed while generating novel narratives and environment states.
Finally, we demonstrate significant data efficiency ($\mathbb{C}1$): starting from 400 human seeds and maintaining an active training pool of $\sim$800 samples, \ours achieves performance comparable to a 17K-sample offline pipeline (using roughly $20\times$ less data) and improves over fixed-data RL in our controlled setting.

\section{Related Work}
\paragraph{Data synthesis for tool use.}
Although high-quality human-annotated benchmarks~\citep{apibank,tbench,tau2bench,acebench,travelplanner,funcbenchgen} provide rigorous evaluation protocols for multi-turn tool use, their limited scale of typically hundreds of instances is insufficient for training reliable agentic policies.
This data scarcity has prompted a shift toward large-scale offline synthesis.
Frameworks such as APIGen-MT~\citep{prabhakar2025apigen}, TOUCAN~\citep{toucan}, and Magnet~\citep{magnet} prioritize corpus scale and structural complexity, often generating millions of trajectories upfront.
While effective for pre-training, these static pipelines remain decoupled from the training process, producing uniform data distributions that cannot track the model's evolving capability boundary ($\mathbb{C}2$).
\looseness=-1

Directed synthesis methods attempt to narrow this focus by targeting specific failure modes.
FunReason-MT~\citep{xu2025funreasonmt} utilizes environment-API graphs and advanced tool-query synthesis to tackle hard query generation, while LoopTool~\citep{looptool} employs a feedback loop to correct algorithmic errors.
Although these represent a clear advance in targeting known weaknesses, they remain offline snapshots that cannot adapt to the shifting capability boundary during active RL training.
\ours addresses this limitation by introducing a reward-driven synthesis loop that operates \emph{during} training, ensuring that the generated data remains at the model's immediate capability boundary.

\paragraph{Online RL and curriculum learning.}
Data efficiency in agentic RL is traditionally addressed through environment simulation or corrective feedback. Simulation-based approaches like ScaleEnv~\citep{scaleenv}, Agent World Model~\citep{agentworldmodel}, and Simia-Agent~\citep{simulatingenv} construct high-fidelity interactive loops to maximize the signal extracted from existing tasks. EnvTuning~\citep{tuneenv} further improves efficiency by orchestrating a four-stage curriculum with actionable environment augmentation. However, these methods are constrained by the fixed diversity of their seed corpora; as the agent improves, the proportion of gradient-informative samples in the static pool shrinks.

Self-play paradigms~\citep{toolr0,selfplayevolve} attempt to resolve this via absolute zero-data generation, where a generator proposes tasks from scratch.
While theoretically appealing, such unconstrained generation frequently struggles with the long-horizon logical chains and interdependent API calls required for multi-turn tool use ($\mathbb{C}3$).
\ours takes a highly complementary approach by using a critically small set of human data as \emph{structural anchors}.
By reusing the inherent rollout variance of GRPO as a boundary detector and grounding synthesis in skill-aligned resampling, \ours enables a targeted, complexity-aligned curriculum that bypasses the high-variance search of generating multi-turn logic from scratch.
Our approach aligns with prioritized data selection methods, such as prioritized experience replay~\citep{schaul2016prioritized}, hard-example mining~\citep{shrivastava2016training}, and competence-based curricula~\citep{platanios2019competence}, which focus learning on the most informative samples. The key distinction is that these classical methods select or re-weight \emph{existing} experiences, whereas \ours generates \emph{new} data at the reward-defined boundary, combining the targeting principle of prioritized replay with the distributional expansion of generative synthesis.

\section{\ours}
\label{sec:methodology}
To resolve multi-turn data scarcity and signal depletion, we introduce \ours, a reward-driven data synthesis framework.
\ours maintains a saturated learning signal through three co-evolving modules.
First, reward-based seed detection identifies high-variance boundary tasks (\S\ref{subsec:detection}).
Second, skill-aligned synthesis generates coherent, structurally-isomorphic variants (\S\ref{subsec:synthesis}).
Third, dynamic replay buffer management tracks the shifting capability boundary (\S\ref{subsec:lifecycle}).

\begin{figure}[!t]
	\centering
	\includegraphics[width=0.95\textwidth]{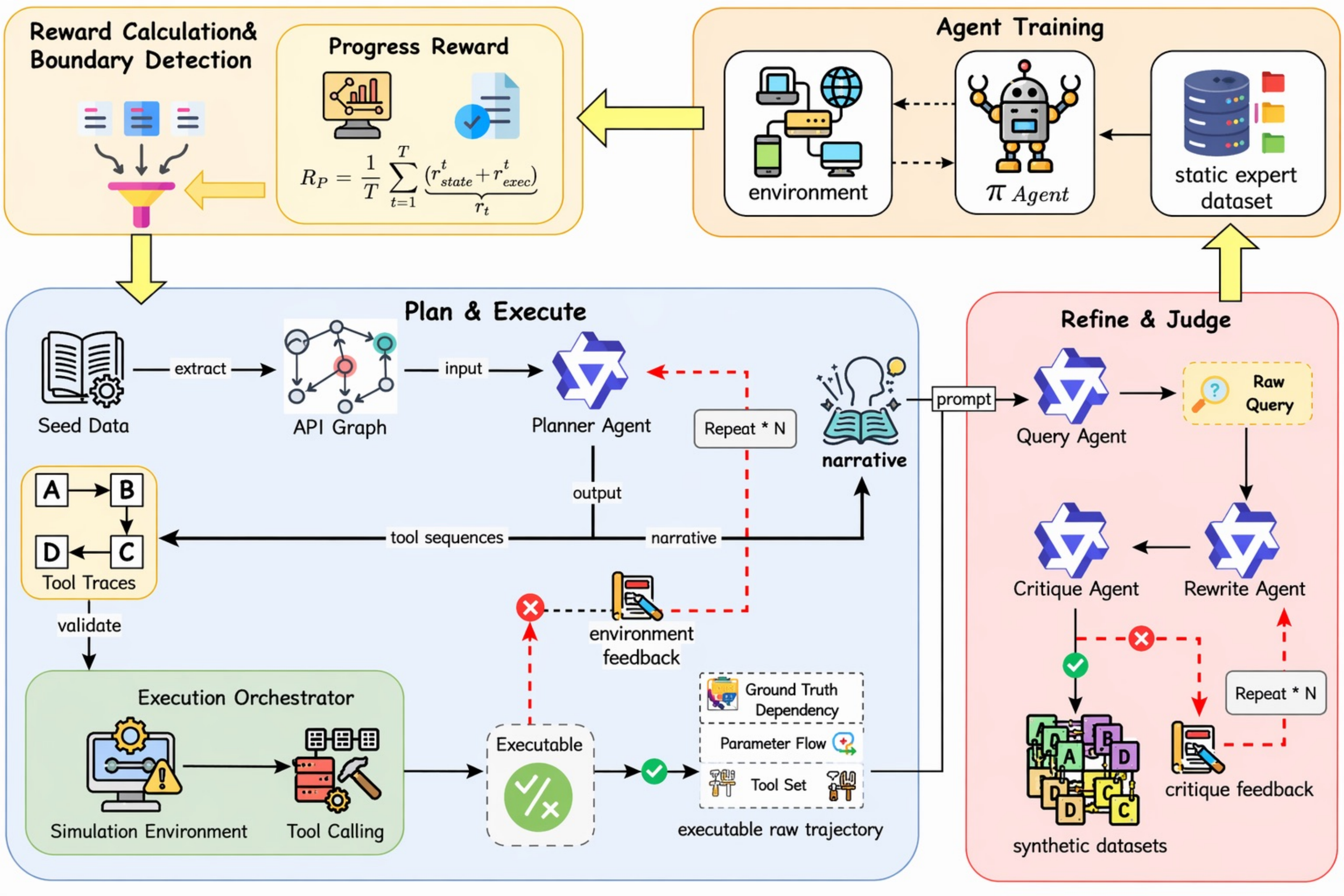}
	\vspace{-1em}
	\caption{
		\textbf{The \ours closed-loop RL-data synthesis architecture.}
		\textbf{(Top)} \emph{Reward Calculation \& Agent Training:} The agent trains on a mixed dataset via GRPO; the Progress Reward ($R_P$) identifies boundary seeds and feeds them back to the data engine.
		\textbf{(Bottom-left)} \emph{Plan \& Execute:} The Planner Agent selects a function sequence from the API graph; the Execution Orchestrator instantiates it on a simulation environment with environment feedback (Repeat $\times N$), producing an executable raw trajectory.
		\textbf{(Bottom-right)} \emph{Refine \& Judge:} A Query Agent converts the trajectory into per-turn natural-language queries; a Rewrite Agent grounds all queries in the Planner's narrative for cross-turn coherence; a Critique Agent validates semantic quality with a feedback loop (Repeat $\times N$), yielding validated synthetic datasets that are injected into training.
	}
	\vspace{-1em}
	\label{fig:architecture}
\end{figure}

\subsection{Problem Formulation and Design Motivation}
\label{subsec:preliminary}
Multi-turn tool use is formalized as a POMDP where an agent resolves interdependent queries through API calls and environment feedback (details in Appendix~\ref{app:pomdp_formulation}). We use GRPO~\citep{shao2024deepseekmath} for optimization, though our variance-based boundary tracking extends to other trajectory-sampling methods like PPO (see Appendix~\ref{app:ppo_extension}).

\paragraph{Reward sparsity and progress reward.}
Because sparse binary rewards fail to assign credit across complex long-horizon trajectories~\citep{feng2025group}, we adopt the progress reward ($R_P \in [0, 1]$) from \citet{tuneenv}: $R_P = \nicefrac{1}{N} \sum_{t=1}^{N} (r_t^{\text{state}} \cdot r_t^{\text{exec}})$. This assigns continuous credit for partial completion, enriching the advantage signal. Note that ground truth is used exclusively for simulation-based reward computation; the policy never observes it.

\paragraph{Design heuristic: variance peaks near the capability boundary.}
In policy gradient methods like GRPO~\citep{shao2024deepseekmath}, the gradient signal density of a sample is governed by its rollout reward variance. While Popoviciu's inequality~\citep{popoviciu1935sur} shows that the variance of a bounded variable is upper-bounded by $\mu(1-\mu)$ (maximizing at $\mu \approx 0.5$), this provides a theoretical upper bound rather than an exact description of the continuous reward variance. We nonetheless validate this as an \emph{empirically supported heuristic}: the continuous progress reward variance peaks near the capability boundary ($\mu \approx 0.5$), as confirmed by the per-task variance analysis in Figure~\ref{fig:dynamic_mechanism} (right). Thus, this heuristic underpins the design of \ours: identifying these boundary samples serves as an effective, cheap proxy for finding the highest concentration of informative gradient signals (see Appendix~\ref{sec:theory} for theoretical details).

\subsection{Reward-Based Seed Data Detection}
\label{subsec:detection}
To operationalize the boundary-targeting heuristic identified in \S\ref{subsec:preliminary}, we use the average Progress Reward $\bar{r}_i$ across $K$ rollouts as a density-based probe to partition the task space $\mathcal{D}$ into three dynamically evolving regions:
\begin{small}
	\begin{equation}
		\mathcal{D}_{\text{mastered}} = \{x_i : \bar{r}_i > \alpha^+\} \,,
		\mathcal{D}_{\text{boundary}} = \{x_i : \alpha^- \leq \bar{r}_i \leq \alpha^+\}\,,
		\mathcal{D}_{\text{hard}} = \{x_i : \bar{r}_i < \alpha^-\} \,,
		\label{eq:partition}
	\end{equation}
\end{small}%
where $\alpha^-$ and $\alpha^+$ are boundary thresholds (set to $0.20$ and $0.85$ respectively, see Appendix~\ref{sec:implementation_details}).
We define the tasks within the intermediate region $x_i \in \mathcal{D}_{\text{boundary}}$ as \textbf{boundary seeds}, as they represent the model's immediate capability boundary and harbor the highest potential for informative gradient signals.
\looseness=-1

\paragraph{Type-quota seed emission.}
At each training step, we select up to $M$ seeds from $\mathcal{D}_{\text{boundary}}$.
To ensure diverse skill coverage, we enforce a per-type quota $M_\tau$ (e.g., base/long-context, missing-function, missing-parameter) such that $\sum_\tau M_\tau = M$.
Within each type, candidates are ranked by the variance-proxy $\phi(\bar{r}_i) = 4\bar{r}_i(1-\bar{r}_i)$ in descending order, prioritizing seeds closest to the capability midpoint ($\bar{r}_i = 0.5$) where gradient signal potential is highest (cf.\ $\mathbb{C}2$).
A temporal exclusion window of $c$ steps prevents redundant sampling, and the selected seeds are asynchronously dispatched to the synthesis pipeline.

\subsection{Skill-Aligned Data Synthesis}
\label{subsec:synthesis}
The synthesis pipeline transforms boundary seeds into novel, structurally valid variants that preserve the informative complexity of the original task (see Appendix~\ref{app:data_examples} for detailed seed-to-variant examples).
Rather than simple paraphrasing, we enforce \emph{structural similarity}: we extract the complexity profile $\Phi(x_{\text{seed}})$ (e.g., the directed acyclic graph of API dependencies and parameter flows) and sample a new task $x' \sim p(\cdot | \Phi(x_{\text{seed}}))$ such that $\Phi(x')$ approximates $\Phi(x_{\text{seed}})$ in dependency depth and API topology.
	This constraint ensures that $x'$ has structural difficulty similar to the seed, placing it near the capability boundary ($\mathbb{C}2$) as validated by the injection reward analysis in Figure~\ref{fig:dynamic_mechanism} (middle), while forcing the model to generalize across novel abstract logic and environment states, preventing overfitting to static execution paths ($\mathbb{C}3$).
	We implement this via a five-stage multi-agent pipeline (see Appendix~\ref{app:agent_prompts} for full prompts):
In Stage I (schema-guided planning, addressing $\mathbb{C}2$ and $\mathbb{C}3$), the planner agent ingests $x_{\text{seed}}$ to design an execution plan $\mathcal{S}'$ and an underlying narrative $\mathcal{N}$, using failure histories ($\mathcal{F}_{\text{avoid}}$) to bypass known execution bottlenecks.
In Stage II (feedback-driven execution, addressing $\mathbb{C}1$), an execution orchestrator instantiates $\mathcal{S}'$ in a simulated environment $\mathcal{C}_0$ to produce a trajectory with ground-truth dependencies.
An error critic applies multi-tier mitigation to find a valid instantiation across $K_{\max}\!=\!3$ attempts, while a query agent maps each turn to natural-language queries without exposing function details.
In Stage III (holistic semantic grounding, addressing $\mathbb{C}3$), a rewrite agent renders the entire multi-turn trajectory simultaneously based on $\mathcal{N}$.
Unlike greedy turn-by-turn generation, this approach anchors all turns to a single goal, resolving semantic disjointedness and ensuring natural conversational flow.
In Stage IV (critique and refinement), variants undergo rule-based checks and LLM scoring via a critique agent. A feedback loop with the rewrite agent prunes logic flaws and fixes phrasing.
Finally, Stage V (optional adversarial augmentation) injects structural exceptions, such as missing tools or blurred parameters, to force clarification turns and improve out-of-distribution robustness.

\subsection{Dynamic Replay Buffer Management}
\label{subsec:lifecycle}
To maintain a high-fidelity training distribution, we implement a dual-control lifecycle that balances \emph{expansion flow} with \emph{stock relevance}.

\paragraph{Expansion flow control: staged injection.}
To prevent instability from abrupt distributional shifts—where a sudden influx of synthetic data destabilizes RL gradients—we adopt a staged injection protocol.
Synthesized variants are initially held in an asynchronous candidate queue and merged into the active replay pool strictly at epoch boundaries.
Specifically, tasks seeded during epoch $n$ are staged and injected at the start of epoch $n{+}1$.
To ensure gradual integration, the per-epoch injection volume is capped at $\beta \cdot |\mathcal{D}_{\text{active}}|$; any excess is deferred to a persistent staging buffer for subsequent epochs.
Upon each injection, the sampling distribution is re-initialized to ensure uniform coverage of the expanded pool.

\paragraph{Active stock management: multi-layer retirement.}
We maintain pool informativeness through a three-layer retirement mechanism that tracks the shifting capability boundary.
The first layer, burn-in filtering, discards new variants with initial rewards below $\epsilon_{\text{trial}}$ to prune tasks beyond the immediate exploration horizon.
The second layer, boundary-drift eviction, removes tasks that have drifted into mastered ($\bar{r}_i > \alpha^+_{\text{retire}}$) or unsolvable ($\bar{r}_i < \alpha^-_{\text{retire}}$) zones after $n_{\text{min}}$ observations.
The third layer applies variance-prioritized pruning: if the pool exceeds $P_{\max}$, samples are retired based on reward variance $\phi(\bar{r}_i) = 4\bar{r}_i(1-\bar{r}_i)$, preserving the highest gradient signals ($\mathbb{C}2$).
We also prune generated variants that remain unsampled for an extended period to prevent stale data accumulation.
This dual-control architecture ensures the replay buffer remains a high-fidelity mirror of the capability boundary while maintaining the stability of the RL optimization loop.
We employ default hyperparameters across all settings to ensure a reliable and fair evaluation; details are provided in Appendix~\ref{app:training_details}.

\section{Experiments}
\label{sec:experiments}

Our experiments are designed to answer five questions:
\textbf{(Q1)} Does boundary-targeted expansion outperform both fixed-data RL and environment augmentation?
\textbf{(Q2)} Does boundary-expanded training generalize to OOD tasks?
\textbf{(Q3)} How does \ours dynamically expand the data space without catastrophic distribution shifts?
\textbf{(Q4)} How data-efficient is \ours compared to large-scale offline synthesis?
\textbf{(Q5)} Which components of \ours contribute most to performance?

\paragraph{Benchmark.}
We evaluate on the multi-turn subset of BFCL V3~\citep{patil2025bfcl}, which includes 800 samples across four balanced splits: \texttt{Base}, \texttt{Missing Function}, \texttt{Missing Parameter}, and \texttt{Long-Context}.
Following~\citet{tuneenv}, we reserve 400 samples (100 per split) for training and use the remaining 400 for held-in evaluation.
For OOD testing, we adopt the BFCL V4 multi-turn tracks, $\tau^2$-bench~\citep{tau2bench}, and the ACEBench Agent split~\citep{acebench} as our held-out test sets.
Detailed benchmark and evaluation details are provided in Appendix~\ref{app:benchmark_details}.

\paragraph{Training configuration.}
We train Qwen3-4B-Instruct using GRPO~\citep{shao2024deepseekmath} ($K\!=\!16$ rollouts) on $8\times$A100 GPUs via a three-stage curriculum (shared across baselines for fair comparison, details in Appendix~\ref{app:training_details}). Stage transitions occur when validation performance plateaus (changing $<$1\% over one full epoch) and gradient norms converge, following the protocol of~\citet{tuneenv}. The protocol isolates syntactic and logical acquisition: (1) \textbf{Stage 1 (Format):} Training on 100 \texttt{Base} samples using a format reward $R_{\text{format}}$ to isolate syntactic acquisition (XML, function names, arguments) before task reasoning (see Appendix~\ref{app:reward_details}); (2) \textbf{Stage 2 (Base Reasoning):} Continuing on the \texttt{Base} split with the Progress Reward $R_P$ to build a stable reasoning anchor without expansion; and (3) \textbf{Stage 3 (Full data + expansion):} Scaling to 400 samples across all splits. For \ours, the synthesis engine targets high-variance boundary tasks, using the Stage 2 foundation to drive generalization. To evaluate cross-model generalizability, we also report results on Qwen2.5-7B-Instruct and Llama-3.1-8B-Instruct.

\paragraph{Data synthesis.}
The \ours synthesis pipeline uses Qwen3-32B deployed via vLLM on a separate $8\times$A100 cluster running asynchronously alongside training. It achieves a seed-to-injection latency of $\sim$1 training step, introducing no idle time into the RL loop (boundary thresholds $\alpha^-\!=\!0.20, \alpha^+\!=\!0.85, P_{\max}\!=\!400$). Full configuration and detailed cost breakdown are provided in Appendix~\ref{app:training_details} and \ref{app:synthesis_cost}.
\looseness=-1

\paragraph{Baselines and references.}
We structure our comparisons into two tiers. \textbf{Tier 1: Controlled RL comparisons} (consistent 400-sample seeds) includes \textsc{Static dataset} (baseline trained only on the fixed seed set without generation), \textsc{EnvTuning}~\citep{tuneenv} (actionable environment enrichment), and \textsc{\ours} (our full dynamic boundary expansion system). \textbf{Tier 2: Data efficiency references} (varying scales) includes \textsc{FunReason-MT-4B}~\citep{xu2025funreasonmt} ($20\times$ more data) and state-of-the-art models like GPT-4o and DeepSeek-V3.2-Exp.

\subsection{Results (Q1): Boundary Expansion vs.\ Fixed-Data RL and Environment Augmentation \looseness=-1}
\begin{table*}[t]
	\small
	\centering
	\resizebox{\textwidth}{!}{
		\begin{tabular}{@{}lcccccc@{}}
			\toprule
			\textbf{Model}
			                              & \textbf{Size}
			                              & \textbf{Overall}
			                              & \textit{Base}
			                              & \begin{tabular}[c]{@{}c@{}}\textit{Miss}\\ \textit{Func}\end{tabular}
			                              & \begin{tabular}[c]{@{}c@{}}\textit{Miss}\\ \textit{Param}\end{tabular}
			                              & \begin{tabular}[c]{@{}c@{}}\textit{Long}\\ \textit{Context}\end{tabular}                                                                                                                                                                                                                   \\
			\midrule
			\rowcolor{gray!15}
			\multicolumn{7}{l}{\emph{\textbf{Reference Models}}}                                                                                                                                                                                                                                                                       \\[1pt]
			GPT-4o-2024-11-20             & -                                                                        & 42.50                                   & 55.50                                   & 34.50                                   & 29.00                                   & 51.00                                   \\
			DeepSeek-V3.2-Exp             & 671B                                                                     & 44.88                                   & 55.00                                   & 49.00                                   & 27.00                                   & 48.50                                   \\
			FunReason-MT-4B (17K offline) & 4B                                                                       & 56.50                                   & 63.00                                   & 53.00                                   & 40.00                                   & 55.00                                   \\
			\midrule
			Qwen2.5-7B-Instruct           & 7B                                                                       & $7.00$                                  & $9.33$                                  & $9.33$                                  & $6.33$                                  & $3.00$                                  \\
			\rowcolor{gray!10}
			\, + Static dataset           & -                                                                        & $36.92$ \improvement{(+29.92)}          & $50.33$ \improvement{(+41.00)}          & $40.33$ \improvement{(+31.00)}          & $29.33$ \improvement{(+23.00)}          & $27.67$ \improvement{(+24.67)}          \\
			\rowcolor{blue!10}
			\, + EnvTuning                & -                                                                        & $37.75$ \improvement{(+30.75)}          & $51.50$ \improvement{(+42.17)}          & $41.00$ \improvement{(+31.67)}          & $30.50$ \improvement{(+24.17)}          & $28.00$ \improvement{(+25.00)}          \\
			\, + \ours(ours)              & -                                                                        & $40.25$ \improvement{(+33.25)}          & $54.00$ \improvement{(+44.67)}          & $43.50$ \improvement{(+34.17)}          & $33.50$ \improvement{(+27.17)}          & $30.00$ \improvement{(+27.00)}          \\
			\hdashline
			Llama-3.1-8B-Instruct         & 8B                                                                       & $5.48$                                  & $6.15$                                  & $6.80$                                  & $3.20$                                  & $5.75$                                  \\
			\rowcolor{gray!10}
			\, + Static dataset           & -                                                                        & $28.25$ \improvement{(+22.77)}          & $28.20$ \improvement{(+22.05)}          & $25.85$ \improvement{(+19.05)}          & $22.15$ \improvement{(+18.95)}          & $36.80$ \improvement{(+31.05)}          \\
			\rowcolor{blue!10}
			\, + EnvTuning                & -                                                                        & $28.38$ \improvement{(+22.90)}          & $28.00$  \improvement{(+21.85)}         & $25.50$ \improvement{(+18.70)}          & $23.00$  \improvement{(+19.80)}         & $37.00$ \improvement{(+31.25)}          \\
			\rowcolor{cyan!8}
			\, + \ours(ours)              & -                                                                        & $30.88$ \improvement{(+25.40)}          & $32.00$ \improvement{(+25.85)}          & $28.00$ \improvement{(+21.20)}          & $24.50$ \improvement{(+21.30)}          & $39.00$ \improvement{(+33.25)}          \\
			\hdashline
			Qwen3-4B-Instruct             & 4B                                                                       & $22.13$                                 & $26.50$                                 & $21.00$                                 & $15.50$                                 & $25.50$                                 \\
			\rowcolor{gray!10}
			\, + Static dataset           & -                                                                        & $50.00$ \improvement{(+27.87)}          & $62.00$ \improvement{(+35.50)}          & $51.00$ \improvement{(+30.00)}          & $35.00$ \improvement{(+19.50)}          & $52.00$ \improvement{(+26.50)}          \\
			\rowcolor{blue!10}
			\, + EnvTuning                & -                                                                        & $50.50$ \improvement{(+28.37)}          & $64.00$ \improvement{(+37.50)}          & $52.00$ \improvement{(+31.00)}          & $35.00$ \improvement{(+19.50)}          & $51.00$ \improvement{(+25.50)}          \\
			\rowcolor{cyan!8}
			\, + \ours(ours)              & -                                                                        & $\mathbf{56.00}$ \improvement{(+33.87)} & $\mathbf{68.00}$ \improvement{(+41.50)} & $\mathbf{59.00}$ \improvement{(+38.00)} & $\mathbf{44.00}$ \improvement{(+28.50)} & $\mathbf{53.00}$ \improvement{(+27.50)} \\
			\hdashline
			\bottomrule
		\end{tabular}
	}
	\vspace{-0.5em}
	\caption{\label{tab:main_results}
		\textbf{In-distribution performance on BFCL V3 multi-turn} (Tier 1: controlled RL comparisons).
		All RL methods share the same 400 training samples and GRPO setup. FunReason-MT-4B (Tier 2) is included for data-scaling reference. For a detailed comparison with 20+ models, see Appendix~\ref{app:full_results}. \improvement{Red text} indicates improvement over the base model.
	}
	\vspace{-1em}
\end{table*}

Table~\ref{tab:main_results} presents the controlled comparison among Tier 1 methods.
All three methods share the same 400 training samples, the same GRPO configuration, and the same progress reward; the only variable is the strategy for addressing gradient signal depletion in Stage 3.

\paragraph{Boundary expansion achieves the best results in our controlled setting.}
\ours attains the highest overall scores across the three model families in our experiments.
On the Qwen3-4B-Instruct base model, \ours improves overall multi-turn performance by \textbf{+33.87\%} (reaching 56.00\%), surpassing both the ``Static dataset'' baseline (50.00\%) and the EnvTuning baseline (50.50\%).
This suggests that dynamically synthesized boundary data provides complementary gradient signals beyond what fixed datasets or enriched environment feedback alone supply.
The gains are consistent across all four sub-splits, indicating that boundary targeting benefits diverse task complexities rather than overfitting to a specific category.
\ours achieves comparable performance to large-scale offline synthesis (FunReason-MT-4B) while using roughly $20\times$ fewer trajectories. Furthermore, it improves over fixed-data RL and environment augmentation in our reported runs under the same 400-sample controlled setup. While \ours also achieves competitive benchmark performance against larger models (e.g., DeepSeek-V3.2), we emphasize that the controlled comparisons within the same 400-sample setup constitute the primary evidence for the mechanism's efficacy.

\paragraph{Data expansion vs.\ environment augmentation: two orthogonal strategies.}
The comparison between \ours and EnvTuning is informative, as both address the same problem (gradient sparsity under data scarcity) but via opposite mechanisms.
EnvTuning enriches the \emph{feedback signal} on existing data by providing corrective hints upon failure, effectively extracting more value from each fixed sample.
\ours instead expands \emph{the data distribution itself} by injecting new variants precisely at the capability boundary.

The observed empirical advantage of \ours (e.g., a +5.50\% absolute lead over EnvTuning on Qwen3-4B) suggests that expanding the data distribution at the capability boundary is more effective than deepening feedback on fixed samples in extreme data-scarce RL settings. The structural diversity of boundary-targeted variants provides gradient signals that an exhausted static pool cannot.
Because \ours achieves this without any structural modification to the training environment's feedback mechanism, it is environment-agnostic and readily applicable to new API domains.
We investigate whether the two strategies are complementary in Appendix~\ref{app:combination}.

\paragraph{OOD generalization (Q2).}
We evaluate on BFCL V4, $\tau^2$-bench, and the ACEBench Agent split to test whether boundary-expanded training yields generalizable reasoning or merely in-distribution pattern matching.
The OOD improvements of \ours over the base model (detailed in Appendix~\ref{app:ood_results}) are consistent with our design hypothesis in Section~\ref{subsec:synthesis}: by preserving API execution plans while heavily randomizing surface variables and environment states, \emph{structural isomorphism} prevents the policy from overfitting to specific textual cues, forcing it to internalize abstract, generalizable multi-turn reasoning patterns.
\subsection{Mechanism validation (Q3) via Data Space Evolution}
\label{subsec:mechanism_validation}

\begin{figure}[t]
	\centering
	\begin{minipage}[b]{0.66\linewidth}
		\includegraphics[width=\linewidth]{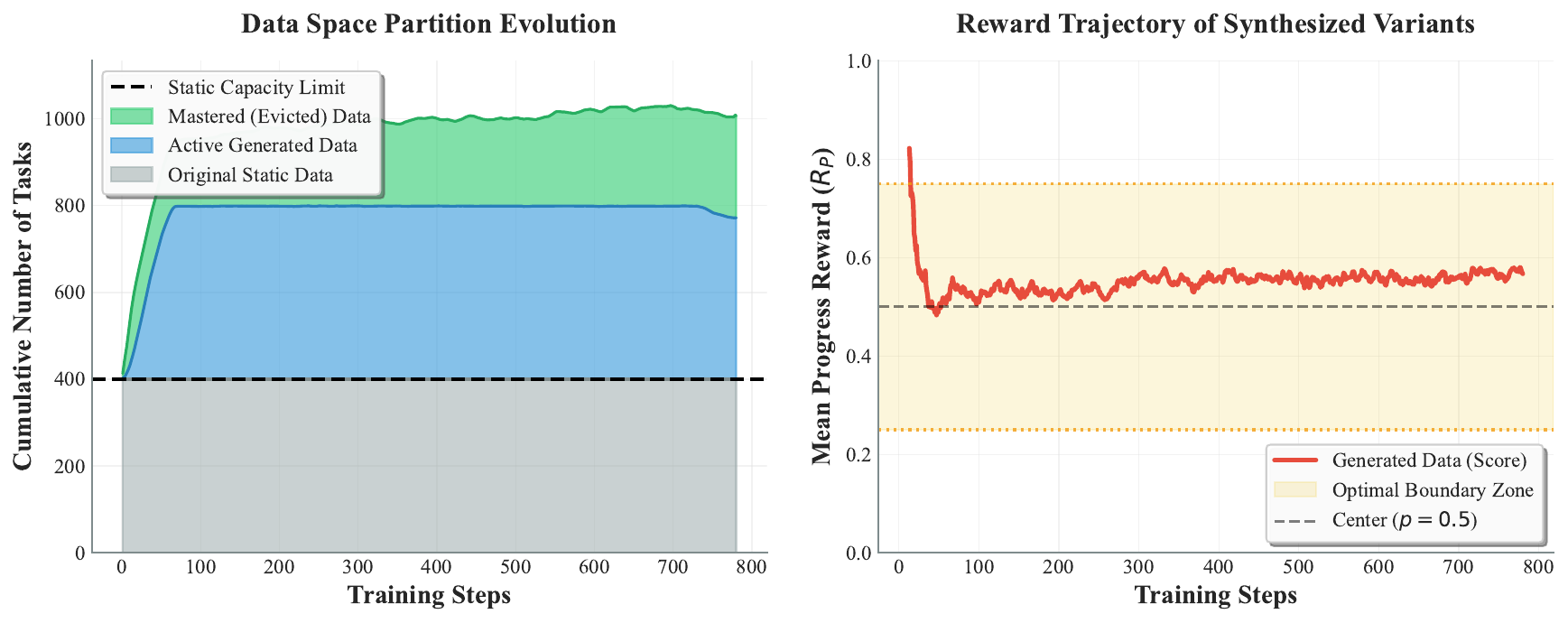}
	\end{minipage}
	\hfill
	\begin{minipage}[b]{0.32\linewidth}
		\includegraphics[width=\linewidth]{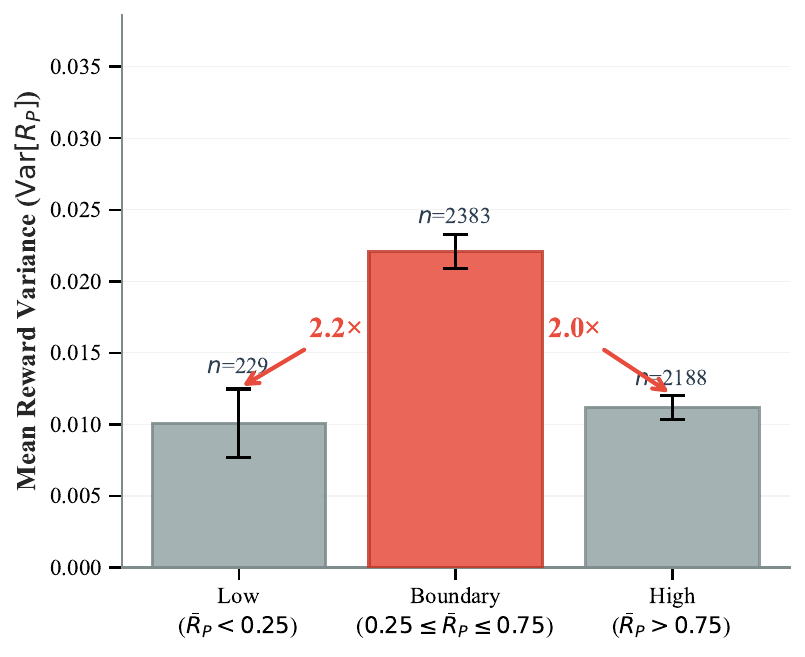}
	\end{minipage}
	\vspace{-1em}
	\caption{\small
		\textbf{The \ours dynamic synthesis mechanism during stage 3 training.}
		\textbf{(Left)} Evolution of the data space partition. \ours breaks the static capacity limit ($400$ tasks) by continuously synthesizing active boundary data and evicting mastered tasks, effectively expanding the training curriculum without exploding memory.
		\textbf{(Middle)} Reward trajectory of newly synthesized variants upon injection. The continuous data generation strictly anchors the mean progress reward within the boundary zone ($[0.25, 0.75]$), confirming successful targeting of the capability frontier.
		\textbf{(Right)} Empirical validation of the variance heuristic. Across 4{,}800 per-task measurements ($K\!=\!16$ rollouts each), rollout reward variance in the boundary zone is $2.0$--$2.2\times$ higher than in the low-reward or high-reward regions, confirming that gradient signal concentrates near the capability boundary.
	}
	\vspace{-1em}
	\label{fig:dynamic_mechanism}
\end{figure}

\paragraph{Breaking the static capacity limit via boundary anchoring.}
Figure~\ref{fig:dynamic_mechanism} illustrates the internal mechanics driving the data efficiency reported in Table~\ref{tab:main_results}. Rather than accumulating data indefinitely, \ours treats the active training pool as a sliding window over the capability space. The left panel shows how the system breaches the 400-task static limit by introducing new variants while retiring mastered ones, generating over 800 unique tasks in total while keeping the active pool bounded by $P_{\max}$. The middle panel confirms that newly synthesized variants land inside the boundary zone ($R_P \in [0.25, 0.75]$). The right panel empirically validates the underlying heuristic: across 4{,}800 per-task measurements, rollout reward variance in the boundary zone is $2.0$--$2.2\times$ higher than in the mastered or too-hard regions, confirming that gradient signal concentrates near the capability boundary and thus alleviating gradient starvation ($\mathbb{C}2$).

\subsection{Data efficiency analysis (Q4): boundary targeting vs.\ blind scaling}
\label{subsec:data_scaling}

Our central claim is that \emph{where} to synthesize data matters more than \emph{how much}.
We evaluate this by (a) scaling the boundary-targeted data pool in \ours, and (b) comparing its efficiency against FunReason-MT~\citep{xu2025funreasonmt}, a state-of-the-art large-scale offline synthesis pipeline.
\looseness=-1

\paragraph{Scaling the maximum generated pool size ($P_{\max}$).}
We vary $P_{\max} \!\in\! \{0, 50, 100, 200, 400\}$ on Qwen3-4B-Instruct while maintaining the standard three-stage curriculum.
\looseness=-1

\begin{figure}[t]
	\centering
	\includegraphics[width=\linewidth]{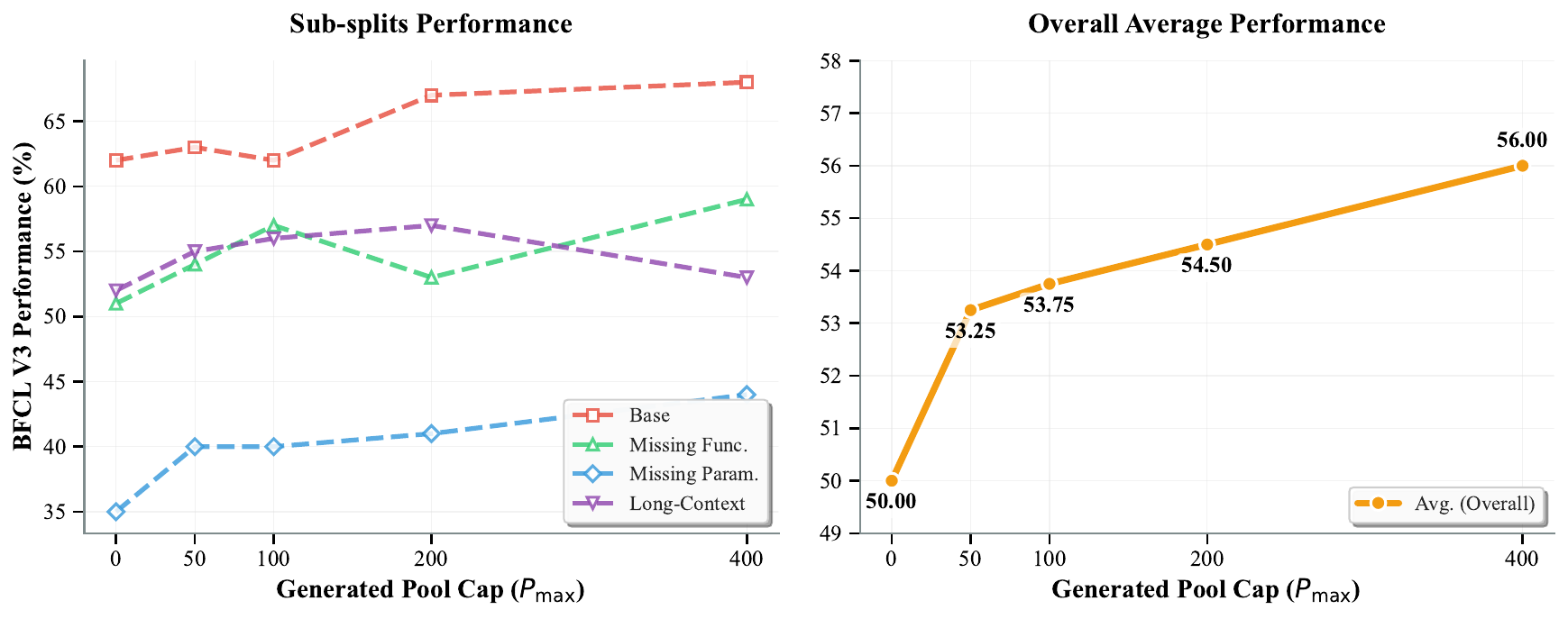}
	\vspace{-2em}
	\caption{\small
		\textbf{Data scaling analysis.}
		Performance on BFCL V3 as a function of the generated pool cap $P_{\max}$.
		$P_{\max}\!=\!0$ corresponds to the static baseline (static pool of 400 items).
		(\textbf{Left}) Performance across sub-splits (Base, Missing Functions, Missing Parameters, Long-Context).
		(\textbf{Right}) The overall average performance (solid orange) steadily improves as $P_{\max}$ increases, with specific average scores annotated.
	}
	\vspace{-1em}
	\label{fig:data_scaling}
\end{figure}

\begin{itemize}[nosep, leftmargin=12pt]
	\item Even $P_{\max}\!=\!50$ ($\sim$12\% expansion) yields a meaningful improvement over the static baseline, demonstrating the high marginal value of a small amount of targeted boundary data.
	\item Performance scales with $P_{\max}$ but exhibits diminishing returns beyond $P_{\max}\!=\!200$, as the boundary region of the 400 original samples becomes fully covered.
	\item This validates $P_{\max}\!=\!400$ as a practical operating point and establishes that boundary data has a significantly higher per-sample training value than uniformly sampled data.
\end{itemize}

\paragraph{Comparison with large-scale offline synthesis.}
To contextualize the data efficiency of boundary targeting, we compare against FunReason-MT-4B~\citep{xu2025funreasonmt} in Table~\ref{tab:main_results}, which was trained on 17K offline trajectories. \ours achieves a highly competitive 56.00\% overall utilizing an active training pool of $\sim$800 samples (400 original seeds + up to $P_{\max}=400$ generated variants), matching FunReason-MT on Long-Context and significantly exceeding it on Missing Functions and Missing Parameters. This represents an $\sim$$20\times$ reduction in data volume to achieve comparable or superior performance, demonstrating that boundary-targeted synthesis is a highly data-efficient alternative to massive offline corpora.

	We further isolate the value of boundary-aware seed selection from the mere benefit of additional data volume via an ablation on random expansion (Table~\ref{tab:ablation}, row ``w/ random seed selection'').

	\subsection{Ablation study (Q5)}
	\label{subsec:ablation}
	We ablate the three pillars of \ours---boundary detection, synthesis pipeline, and lifecycle management---to quantify their individual contributions (Table~\ref{tab:ablation}). All ablations use Qwen3-4B-Instruct with $P_{\max}\!=\!400$ under the same three-stage curriculum. Detailed ablation configuration descriptions are provided in Appendix~\ref{app:ablation_details}.

	\begin{table}[!t]
		\centering
		\caption{\small
			\textbf{Ablation study on BFCL V3.}
			Each row modifies one component of \ours while keeping the rest intact.
			$\Delta$: change relative to the full system.
		}
		\resizebox{0.8\linewidth}{!}{
			\begin{tabular}{l ccccc c}
				\toprule
				\multirow{2}{*}{\textbf{Configuration}}        & \multicolumn{5}{c}{\textbf{BFCL V3 Multi Turn}} & \multirow{2}{*}{$\Delta$\textbf{Avg.}}                                                                               \\
				\cmidrule(lr){2-6}
				                                               & \textbf{Avg.}                                   & \textbf{Base}                          & \textbf{M. Func} & \textbf{M. Param} & \textbf{L. Ctxt}                     \\
				\midrule
				\rowcolor{cyan!8}
				\textbf{\ours (full system)}                   & \textbf{56.00}                                  & \textbf{68.00}                         & \textbf{59.00}   & \textbf{44.00}    & \textbf{53.00}   & ---               \\
				\midrule
				\multicolumn{7}{l}{\emph{\textbf{(a) Boundary detection}}}                                                                                                                                                              \\[1pt]
				\rowcolor{gray!10}
				\quad w/ random seed selection                 & 51.25                                           & 63.50                                  & 52.50            & 37.00             & 52.00            & $\downarrow$ 4.75 \\
				\quad w/ binary acc instead of progress reward & 52.75                                           & 65.00                                  & 55.00            & 36.50             & 54.50            & $\downarrow$ 3.25 \\
				\midrule
				\multicolumn{7}{l}{\emph{\textbf{(b) Synthesis pipeline}}}                                                                                                                                                              \\[1pt]
				\rowcolor{gray!10}
				\quad w/o coherence rewrite                    & 50.87                                           & 63.00                                  & 52.00            & 36.00             & 52.50            & $\downarrow$ 5.13 \\
				\quad w/o narrative planning                   & 52.37                                           & 64.50                                  & 54.00            & 42.00             & 49.00            & $\downarrow$ 3.63 \\
				\rowcolor{gray!10}
				\quad w/o feedback loop (blind retry)          & 53.87                                           & 66.00                                  & 57.00            & 40.50             & 52.00            & $\downarrow$ 2.13 \\
				\midrule
				\multicolumn{7}{l}{\emph{\textbf{(c) Lifecycle management}}}                                                                                                                                                            \\[1pt]
				\rowcolor{gray!10}
				\quad w/o retirement mechanism                 & 52.62                                           & 64.50                                  & 55.00            & 39.00             & 52.00            & $\downarrow$ 3.38 \\
				\quad w/ static pool (no dynamic refresh)      & 53.12                                           & 65.50                                  & 56.00            & 39.50             & 51.50            & $\downarrow$ 2.88 \\
				\bottomrule
			\end{tabular}
		}
		\label{tab:ablation}
	\end{table}

	\paragraph{Key findings.}
	\textbf{(a)} Removing coherence rewrite ($-5.13\%$) causes the largest overall drop, collapsing the Quality Judge pass rate from $\sim$63\% to $\sim$12\% and drastically reducing usable variants. Among boundary detection ablations, random seed selection ($-4.75\%$) confirms that boundary targeting---not merely additional data---drives improvement. Replacing progress reward with binary accuracy ($-3.25\%$) further validates the importance of continuous credit for boundary identification.
	\textbf{(b)} The synthesis pipeline is robust to backbone choice: replacing Qwen3-32B with GLM-4.5-Air yields only $-0.75\%$ (see Appendix~\ref{app:synthesis_llm_robustness}).
	\textbf{(c)} Disabling retirement ($-3.38\%$) confirms that mastered data accumulation dilutes gradient signal; continuous pool refresh is necessary.

	\section{Conclusion}
Training multi-turn tool-use agents via RL faces a tension between data scarcity and signal relevance: static datasets lose informativeness as the agent improves.
This paper introduces \ours, a framework that recasts this bottleneck as a dynamic curriculum design problem.
Because progress reward variance peaks near the agent's shifting capability boundary, \ours reuses this signal as a practical heuristic for boundary detection to sustain informative policy gradients throughout training.
Combined with a narrative-driven, structurally isomorphic synthesis pipeline, our approach achieves competitive in-distribution and OOD improvements using roughly $20\times$ less data than massive offline pipelines.
These results suggest that boundary-targeted synthesis achieves higher per-sample training value than uniformly scaled static corpora, though at additional synthesis compute cost.

\textbf{Limitations \& Future Work.} While \ours provides a highly efficient curriculum, its current synthesis pipeline relies on deterministic simulation environments (implemented via executable Python objects) to verify execution correctness and provide feedback. Adapting this framework to inherently opaque environments or remote Model Context Protocol (MCP) servers remains an area for refinement. Future work will explore extending our simulation abstraction to robustly wrap and interact with stateful MCP endpoints, allowing the synthesis engine to safely capture input-observation dynamics without direct access to the underlying internal state. In addition, investigating multi-backbone synthesis ensembles to inject diverse structural priors at the boundary presents a promising direction for scaling agentic capabilities.
\looseness=-1

\newpage
\bibliography{resources/reference}
\bibliographystyle{plainnat}

\newpage
\appendix

\section{Formulation of the Sequential Multi-turn Decision Process}
\label{app:pomdp_formulation}

We formalize the multi-turn tool-use problem within the framework of a Partially Observable Markov Decision Process (POMDP)~\citep{williams2007reinforcement}.
In this context, a single episode represents an entire user task, comprising a series of predefined sequential instructions, referred to as turns.

Formally, we represent the sequence of user instructions as $q_1, q_2, \dots, q_N$.
The episode initiates with the initial observation $o_0$, which encapsulates the first instruction $q_1$ alongside the documentation of accessible tools.
To fulfill this instruction, a sequence of operations is executed. At each timestep $t$, an action $a_t$ is sampled from a predefined action space $\mathcal{A}$ based on the policy $\pi_\theta(a_t|o_t)$. This action space consists of two primary categories:

\begin{itemize}[leftmargin=12pt, nosep]
	\item \textbf{Tool Invocation ($a_t^{\text{tool}}$):} A formatted request to interact with one or multiple external APIs to acquire necessary context (e.g., \texttt{<tool\_call>...</tool\_call>}). Following execution, the environment yields a new observation containing the execution results.
	\item \textbf{Task Resolution ($a_t^{\text{answer}}$):} A conversational response directed at the user (e.g., \texttt{<answer>...</answer>}). Emitting this action signifies the completion of the current sub-task, triggering the environment to provide the subsequent user instruction.
\end{itemize}

For any given turn $i$, the interaction proceeds as an intermediate trajectory of tool invocations, culminating when a task resolution is output.
Once $a_t^{\text{answer}}$ is generated, the environment advances the state by embedding the next instruction $q_{i+1}$ into the subsequent observation $o_{t+1}$. This iterative process continues until all $N$ instructions are completed.

A complete episode yields a full trajectory $\tau = (o_0, a_0, o_1, a_1, \dots, o_T)$, which terminates at timestep $T$ when the final response for the ultimate instruction $q_N$ is issued.
Importantly, it is only at this terminal step $T$ that a \textit{sparse, binary reward} $R_T \in \{0, 1\}$ is assigned, reflecting the overall success or failure of the entire task.
Such delayed and sparse feedback makes credit assignment and exploration during reinforcement learning particularly hard, a common obstacle in long-horizon scenarios.

Therefore, the primary objective is to optimize the policy parameters $\theta$ to maximize the expected terminal reward:
\begin{equation} \label{eq:objective}
	J(\theta) = \mathbb{E}_{\tau \sim \pi_\theta, P}[R_T]
\end{equation}

\section{System Implementation and Hyperparameters}
\label{sec:implementation_details}

In this section, we detail the specific hyperparameters, engineering optimizations, and fault-tolerance mechanisms used in our system implementation.

\subsection{Hyperparameter Specifications}

\textbf{Dataset Partitioning Thresholds.}
As introduced in Section 3.1, the dataset is partitioned into three zones. Empirically, we set the boundary thresholds as follows:
\begin{itemize}[leftmargin=12pt, nosep]
	\item \textbf{Boundary Zone}: $\alpha^- = 0.20$ and $\alpha^+ = 0.85$. This captures tasks where the model exhibits partial but incomplete mastery, maximizing gradient variance.
	\item \textbf{Mastered Zone}: $\alpha^+_{\text{retire}} = 0.95$.
	\item \textbf{Too Hard Zone}: $\alpha^-_{\text{retire}} = 0.20$.
\end{itemize}

\textbf{Injection and Capacity Limits.}
To prevent distribution shock (Section 3.3), the per-epoch injection volume is strictly capped at $20\%$ of the active pool size. Additionally, the active dataset is bounded by a maximum capacity $P_{\max} = 400$ generated items, triggering the priority-based eviction strategy when exceeded. The original seed dataset is strictly preserved and exempt from retirement.

\subsection{Data Injection Latency Analysis}

Variants generated during training step $t$ may not enter the dataset until step $t + \Delta$.
Empirically, we observe a seed-to-variant synthesis delay averaging 1 step, and a total staging delay (awaiting the next epoch boundary) averaging 13 steps (roughly 1--2 full epochs).
Within this asynchronous window, a data point's pass@16 changes by less than $0.10$ on average, ensuring it remains in its originally classified zone.
As a result, freshly generated variants remain highly relevant to the evolving model capability upon injection, validating the asynchronous design.

\subsection{System Engineering and Concurrency}

\textbf{Correctness Guarantees: Simulation Environment Execution as Oracle.}
The Synthesis Simulation Environment is implemented as a faithful replica of the training simulation environment.
If a variant executes successfully on the Simulation Environment during generation, it is mathematically guaranteed to execute without environment-level errors during training.
This architectural choice eliminates the need for costly and error-prone LLM-based correctness judgments, replacing them with deterministic, reproducible data validation.

\textbf{Concurrency and Isolation.}
To prevent resource contention during online synthesis, each training environment instance maintains strictly isolated model and environment states.
The asynchronous background generation module operates in a fully decoupled process space, utilizing filesystem-based inter-process communication with file-level locking.
This design prevents shared-memory race conditions and allows the generation daemon to scale independently of the primary reinforcement learning loop.

\textbf{Prompt Engineering: \texttt{<reason>} vs. \texttt{<think>}.}
To prevent modern reasoning models (e.g., Qwen3, QwQ) from defaulting to unconstrained native thinking modes that disrupt structured parsing, our LLM agents use custom \texttt{<reason>} tags for Chain-of-Thought generation rather than standard reasoning tokens (e.g., \texttt{<think>}). Because \texttt{<reason>} is not mapped to a special token in the tokenizer's vocabulary, it functions purely as a structural prompt directive, ensuring stable and parseable JSON/XML outputs.

\subsection{Crash Recovery and Fault Tolerance}

To support reliable, long-running RL experiments, the system is designed with full fault tolerance.
PromptTracker state is serialized to \texttt{tracker.json} after each epoch boundary.
Generated variants are written to immutable append-only logs (e.g., \texttt{expanded\_epoch\_*.jsonl}).
On daemon or trainer restart, the SeedManager reconstructs the full synthesized dataset and training state by:
\begin{enumerate}[leftmargin=12pt, nosep]
	\item Loading the latest \texttt{tracker.json} to recover PromptTracker historical windows.
	\item Replaying all \texttt{expanded\_*.jsonl} files to rebuild the list of generated data.
	\item Deterministically recomputing the data retirement logic from stored metrics.
\end{enumerate}
This guarantees zero data loss across restart boundaries and lets the curriculum resume where it was interrupted.

\subsection{Motivating Analysis: Gradient Variance at the Capability Boundary}
\label{sec:theory}

We present the analysis motivating our boundary-targeting design heuristic.
Specifically, we establish the rationale for preferentially synthesizing new data from tasks where the model's average progress reward is near the midpoint of its range ($\mu \approx 0.5$).
This heuristic is grounded in the relationship between reward variance and gradient signal strength.

In GRPO, for a given prompt $x$, the model generates $K$ rollouts $y_1, \dots, y_K$.
The corresponding binary rewards (success = 1, failure = 0) are denoted as $r_i \in \{0, 1\}$. While the Progress Reward ($R_P$) used in our training is a multi-valued dense signal rather than strictly binary, Popoviciu's inequality states that the variance of any bounded variable $X \in [0,1]$ is upper-bounded by $\mu(1-\mu)$. This bound motivates our design heuristic: the gradient signal potential is highest near $\mu = 0.5$, though the actual variance depends on the full reward distribution and may not saturate this bound.
The policy gradient is estimated as:
\begin{equation}
	\nabla_\theta J(\theta) \approx \frac{1}{K} \sum_{i=1}^K \hat{A}_i \nabla_\theta \log \pi_\theta(y_i|x)
\end{equation}
where $\hat{A}_i$ is the z-score normalized advantage: $\hat{A}_i = \frac{r_i - \mu_x}{\sigma_x + \epsilon}$.
Here, $\mu_x$ is the empirical success rate (pass@$K$, denoted as $p$), and $\sigma_x = \sqrt{p(1-p)}$ is the standard deviation.

\textbf{Variance maximization in the binary case.}
For binary rewards, the normalized advantage for successful ($r_i=1$) and failed ($r_i=0$) rollouts can be exactly computed as $\hat{A}^+ = \sqrt{\frac{1-p}{p}}$ and $\hat{A}^- = -\sqrt{\frac{p}{1-p}}$, respectively.
The total magnitude of the gradient signal (signal density) is proportional to the variance of the Bernoulli distribution:
\begin{equation}
	\text{Signal Density} \propto p(1-p)
\end{equation}
When a task is too hard ($p \to 0$) or already mastered ($p \to 1$), the signal density vanishes because all rollouts yield identical rewards, resulting in zeroed-out advantages.
For binary rewards, this signal density is maximized at $p = 0.5$, where the model generates a balanced mix of successes and failures, allowing GRPO to clearly contrast correct and incorrect behaviors. For the continuous Progress Reward $R_P \in [0,1]$, the Popoviciu bound provides an analogous (though not necessarily tight) guideline: targeting $\mu \approx 0.5$ maximizes the \emph{potential} for high variance, making it a practical proxy for the learning frontier.
Because GRPO inherently generates these $K$ rollouts during the standard forward pass to compute the group advantage, we can identify these high-variance $\mu \approx 0.5$ tasks at \textbf{zero additional inference cost}.

\section{Agent Prompt Design}
\label{app:agent_prompts}
This section provides the complete prompt templates used by the multi-agent synthesis pipeline. Each agent is implemented as an LLM-based component with carefully designed system and user prompts to ensure high-quality data generation.

\subsection{Planner Agent}
\label{app:planner_prompt}

The Planner Agent is responsible for generating a novel function call sequence that matches the structural complexity of the boundary seed task while introducing variation in the specific functions and scenario.

\begin{tcolorbox}[
		breakable,
		colback=white,
		colframe=purple!80,
		boxrule=1.5pt,
		arc=4pt,
		left=8pt,
		right=8pt,
		top=6pt,
		bottom=6pt,
		width=0.96\textwidth,
		title={\textbf{Planner Agent User Prompt}},
		fonttitle=\small\bfseries,
		coltitle=white,
		colbacktitle=purple!80
	]
	\small

	\begin{tcolorbox}[
			colback=gray!10,
			colframe=gray!60,
			boxrule=1pt,
			arc=3pt,
			left=6pt,
			right=6pt,
			top=4pt,
			bottom=4pt
		]
		\textbf{\# Task}\\[0.3em]
		You are a function call planner for a multi-turn tool-calling benchmark.

		\vspace{0.3em}
		You are given a seed task with its user queries and ground truth function call sequence.
		Your goal is to select a function sequence from the available functions that tests
		SIMILAR capabilities as the seed -- such as parameter extraction, multi-step reasoning,
		cross-turn dependency, etc. -- but using potentially DIFFERENT functions.

		\vspace{0.3em}
		You may use both HIGH-LEVEL functions (which the system will automatically decompose
		into multiple bottom-level calls) and BOTTOM-LEVEL functions (executed directly).

		\vspace{0.3em}
		\textbf{\# Seed Task}\\
		Classes: \texttt{\{classes\_str\}}\\
		User queries: \texttt{\{queries\_text\}}\\
		Ground truth function sequence: \texttt{\{gt\_summary\}}

		\vspace{0.3em}
		\textbf{\# Available Functions}\\
		\texttt{\{func\_list\}}
	\end{tcolorbox}

	\vspace{0.3em}

	\begin{tcolorbox}[
			colback=orange!8,
			colframe=orange!60,
			boxrule=1pt,
			arc=3pt,
			left=6pt,
			right=6pt,
			top=4pt,
			bottom=4pt
		]
		\textbf{\# Guidelines}
		\begin{enumerate}[leftmargin=1.5em, itemsep=0.2em]
			\item Select functions that test similar skills to the seed task (e.g., if the seed requires multi-step parameter passing, your plan should also require it)
			\item HIGH-LEVEL functions are preferred when available -- they produce richer, multi-step call sequences after decomposition
			\item Each turn should have 1-3 functions from the SAME class
			\item For multi-class seeds, alternate classes across turns (e.g., Turn 1: ClassA, Turn 2: ClassB, Turn 3: ClassA)
			\item Output 2-5 turns total
			\item Ensure the sequence is logically coherent (e.g., authenticate before posting, fill fuel before driving)
		\end{enumerate}
	\end{tcolorbox}

	\vspace{0.3em}

	\begin{tcolorbox}[
			colback=toolresponsecolor!8,
			colframe=toolresponsecolor!60,
			boxrule=1pt,
			arc=3pt,
			left=6pt,
			right=6pt,
			top=4pt,
			bottom=4pt
		]
		\textbf{\# Output Format}\\[0.3em]
		First, analyze the seed task and plan your approach inside \texttt{<reason></reason>} tags.
		Then output a brief narrative scenario (2-3 sentences) inside \texttt{<narrative>} tags.
		Finally output each turn inside a \texttt{<turn>} tag.

		\vspace{0.3em}
		IMPORTANT: Output 2-5 turns total. Do NOT output more than 5 turns.\\
		IMPORTANT: ONLY use function names that appear in the Available Functions list above.

		\vspace{0.3em}
		\textbf{Format:}\\
		\texttt{<reason> Your step-by-step analysis... </reason>}\\
		\texttt{<narrative> A user named [name] wants to [goal]... </narrative>}\\
		\texttt{<turn> ClassName: func1, func2, func3 </turn>}\\
		\texttt{<turn> ClassName: func4 </turn>}

		\vspace{0.3em}
		\textbf{Example 1} (single class -- VehicleControlAPI):\\
		\texttt{<turn> VehicleControlAPI: fillFuelTankWithLiter </turn>}\\
		\texttt{<turn> VehicleControlAPI: activateParkingBrake, pressBrakePedal, startEngine </turn>}\\
		\texttt{<turn> VehicleControlAPI: estimate\_drive\_feasibility\_between\_city </turn>}

		\vspace{0.3em}
		\textbf{Example 2} (multi-class -- TwitterAPI + TravelAPI):\\
		\texttt{<turn> TravelAPI: get\_flight\_cost\_from\_cityA\_to\_cityB </turn>}\\
		\texttt{<turn> TravelAPI: book\_flight </turn>}\\
		\texttt{<turn> TwitterAPI: post\_tweet\_with\_my\_acount </turn>}

		\vspace{0.3em}
		\textbf{Example 3} (multi-class -- GorillaFileSystem + MathAPI):\\
		\texttt{<turn> GorillaFileSystem: find\_cat </turn>}\\
		\texttt{<turn> GorillaFileSystem: grep\_function\_name </turn>}\\
		\texttt{<turn> MathAPI: mean </turn>}

		\vspace{0.3em}
		\textbf{Example 4} (single class -- TradingBot, using high-level functions):\\
		\texttt{<turn> TradingBot: update\_market\_status\_with\_current\_time </turn>}\\
		\texttt{<turn> TradingBot: add\_to\_watchlist\_with\_company\_name </turn>}\\
		\texttt{<turn> TradingBot: get\_stock\_info\_with\_company\_name </turn>}\\
		\texttt{<turn> TradingBot: place\_order\_with\_market\_price </turn>}

		\vspace{0.3em}
		Now generate a function plan for the seed task above.
	\end{tcolorbox}

\end{tcolorbox}

\subsection{Config Patch Agent (Error Critic)}
\label{app:config_patch_prompt}

The Config Patch Agent analyzes execution failures and generates environment configuration patches to resolve state conflicts. This agent implements the Environment Patches mechanism described in Stage II of the synthesis pipeline.

\begin{tcolorbox}[
		breakable,
		colback=white,
		colframe=purple!80,
		boxrule=1.5pt,
		arc=4pt,
		left=8pt,
		right=8pt,
		top=6pt,
		bottom=6pt,
		width=0.96\textwidth,
		title={\textbf{Config Patch Agent System Prompt}},
		fonttitle=\small\bfseries,
		coltitle=white,
		colbacktitle=purple!80
	]
	\small

	\begin{tcolorbox}[
			colback=gray!10,
			colframe=gray!60,
			boxrule=1pt,
			arc=3pt,
			left=6pt,
			right=6pt,
			top=4pt,
			bottom=4pt
		]
		\textbf{System Prompt:}\\[0.3em]
		You are an expert at diagnosing configuration issues in function-calling systems.
		You analyze why a function failed with a given config, and suggest minimal fixes.
	\end{tcolorbox}

\end{tcolorbox}

\begin{tcolorbox}[
		breakable,
		colback=white,
		colframe=purple!80,
		boxrule=1.5pt,
		arc=4pt,
		left=8pt,
		right=8pt,
		top=6pt,
		bottom=6pt,
		width=0.96\textwidth,
		title={\textbf{Config Patch Agent User Prompt}},
		fonttitle=\small\bfseries,
		coltitle=white,
		colbacktitle=purple!80
	]
	\small

	\begin{tcolorbox}[
			colback=gray!10,
			colframe=gray!60,
			boxrule=1pt,
			arc=3pt,
			left=6pt,
			right=6pt,
			top=4pt,
			bottom=4pt
		]
		\textbf{Input:}\\[0.3em]
		A data generation pipeline failed with this error:\\
		\hspace{1em}Error type: \texttt{\{error\_type\}}\\
		\hspace{1em}Failed function: \texttt{\{error\_function\}}\\
		\hspace{1em}Detail: \texttt{\{error\_detail\}}

		\vspace{0.3em}
		The initial\_config used was:\\
		\texttt{\{config\_str\}}
	\end{tcolorbox}

	\vspace{0.3em}

	\begin{tcolorbox}[
			colback=orange!8,
			colframe=orange!60,
			boxrule=1pt,
			arc=3pt,
			left=6pt,
			right=6pt,
			top=4pt,
			bottom=4pt
		]
		\textbf{Task:}\\[0.2em]
		Analyze WHY this function failed with this config, and suggest the MINIMUM config change(s) to fix the issue.

		\vspace{0.3em}
		\textbf{Common Fixes:}
		\begin{itemize}[leftmargin=1.5em, itemsep=0.1em]
			\item market\_status ``Closed'' $\rightarrow$ ``Open'' for trading functions
			\item authenticated: false $\rightarrow$ true for functions requiring login
			\item Add pending orders for cancel\_order
			\item Ensure sufficient balance for transactions
			\item fuelLevel too low $\rightarrow$ increase for driving functions
		\end{itemize}
	\end{tcolorbox}

	\vspace{0.3em}

	\begin{tcolorbox}[
			colback=toolresponsecolor!8,
			colframe=toolresponsecolor!60,
			boxrule=1pt,
			arc=3pt,
			left=6pt,
			right=6pt,
			top=4pt,
			bottom=4pt
		]
		\textbf{\# Output Format}\\[0.3em]
		First, analyze the error inside \texttt{<reason></reason>} tags.
		Then output each field change as a \texttt{<patch>} block:

		\vspace{0.3em}
		\texttt{<reason> Your analysis... </reason>}\\
		\texttt{<patch>}\\
		\hspace{1em}\texttt{<class>ClassName</class>}\\
		\hspace{1em}\texttt{<field>field\_name</field>}\\
		\hspace{1em}\texttt{<value>new\_value</value>}\\
		\texttt{</patch>}

		\vspace{0.3em}
		For nested fields, use dot notation: e.g., \texttt{orders.12345.status}

		\vspace{0.3em}
		Now analyze the error above and output the necessary patches.
	\end{tcolorbox}

\end{tcolorbox}

\subsection{Coherence Rewrite Agent}
\label{app:coherence_rewrite_prompt}

The Coherence Rewrite Agent performs holistic semantic grounding (Stage III) by generating natural user queries that are semantically coherent across all turns, guided by the latent narrative.

\begin{tcolorbox}[
		breakable,
		colback=white,
		colframe=purple!80,
		boxrule=1.5pt,
		arc=4pt,
		left=8pt,
		right=8pt,
		top=6pt,
		bottom=6pt,
		width=0.96\textwidth,
		title={\textbf{Coherence Rewrite Agent System Prompt}},
		fonttitle=\small\bfseries,
		coltitle=white,
		colbacktitle=purple!80
	]
	\small

	\begin{tcolorbox}[
			colback=gray!10,
			colframe=gray!60,
			boxrule=1pt,
			arc=3pt,
			left=6pt,
			right=6pt,
			top=4pt,
			bottom=4pt
		]
		\textbf{System Prompt:}\\[0.3em]
		You are a helpful assistant that writes natural user queries for function-calling conversations.
	\end{tcolorbox}

\end{tcolorbox}

\begin{tcolorbox}[
		breakable,
		colback=white,
		colframe=purple!80,
		boxrule=1.5pt,
		arc=4pt,
		left=8pt,
		right=8pt,
		top=6pt,
		bottom=6pt,
		width=0.96\textwidth,
		title={\textbf{Coherence Rewrite Agent User Prompt}},
		fonttitle=\small\bfseries,
		coltitle=white,
		colbacktitle=purple!80
	]
	\small

	\begin{tcolorbox}[
			colback=gray!10,
			colframe=gray!60,
			boxrule=1pt,
			arc=3pt,
			left=6pt,
			right=6pt,
			top=4pt,
			bottom=4pt
		]
		\textbf{\# Task}\\[0.3em]
		Rewrite user queries for a multi-turn function-calling conversation.

		\vspace{0.3em}
		\textbf{\# Scenario}\\
		\texttt{\{narrative\}}

		\vspace{0.3em}
		\textbf{\# Ground Truth Function Calls Per Turn}\\
		\texttt{\{turns\_for\_rewrite\}}
	\end{tcolorbox}

	\vspace{0.3em}

	\begin{tcolorbox}[
			colback=orange!8,
			colframe=orange!60,
			boxrule=1pt,
			arc=3pt,
			left=6pt,
			right=6pt,
			top=4pt,
			bottom=4pt
		]
		\textbf{\# Instructions}\\[0.2em]
		Generate one natural user query for EACH turn that:
		\begin{enumerate}[leftmargin=1.5em, itemsep=0.2em]
			\item Accurately describes what the GT function calls do (without mentioning function names)
			\item Sounds like a real user talking to an AI assistant
			\item Is semantically coherent with the narrative scenario
			\item Does NOT contain special characters that could cause parsing issues
			\item For turns with multiple function calls, the query should naturally imply all of them
		\end{enumerate}
	\end{tcolorbox}

	\vspace{0.3em}

	\begin{tcolorbox}[
			colback=toolresponsecolor!8,
			colframe=toolresponsecolor!60,
			boxrule=1pt,
			arc=3pt,
			left=6pt,
			right=6pt,
			top=4pt,
			bottom=4pt
		]
		\textbf{\# Output Format}\\[0.3em]
		Output each query inside \texttt{<query>} tags, one per turn:\\
		\texttt{<query>Turn 1 user query here</query>}\\
		\texttt{<query>Turn 2 user query here</query>}\\
		\ldots

		\vspace{0.3em}
		IMPORTANT: Output exactly \texttt{\{num\_turns\}} queries, one for each turn.
	\end{tcolorbox}

\end{tcolorbox}

\subsection{Quality Judge Agent}
\label{app:quality_judge_prompt}

The Quality Judge validates synthesized trajectories against strict quality criteria to ensure data integrity before injection into the training pool.

\begin{tcolorbox}[
		breakable,
		colback=white,
		colframe=purple!80,
		boxrule=1.5pt,
		arc=4pt,
		left=8pt,
		right=8pt,
		top=6pt,
		bottom=6pt,
		width=0.96\textwidth,
		title={\textbf{Quality Judge Agent System Prompt}},
		fonttitle=\small\bfseries,
		coltitle=white,
		colbacktitle=purple!80
	]
	\small

	\begin{tcolorbox}[
			colback=gray!10,
			colframe=gray!60,
			boxrule=1pt,
			arc=3pt,
			left=6pt,
			right=6pt,
			top=4pt,
			bottom=4pt
		]
		\textbf{System Prompt:}\\[0.3em]
		You are a strict quality judge for multi-turn function-calling data.
		You evaluate whether a generated sample meets quality standards.
	\end{tcolorbox}

\end{tcolorbox}

\begin{tcolorbox}[
		breakable,
		colback=white,
		colframe=purple!80,
		boxrule=1.5pt,
		arc=4pt,
		left=8pt,
		right=8pt,
		top=6pt,
		bottom=6pt,
		width=0.96\textwidth,
		title={\textbf{Quality Judge Agent User Prompt}},
		fonttitle=\small\bfseries,
		coltitle=white,
		colbacktitle=purple!80
	]
	\small

	\begin{tcolorbox}[
			colback=gray!10,
			colframe=gray!60,
			boxrule=1pt,
			arc=3pt,
			left=6pt,
			right=6pt,
			top=4pt,
			bottom=4pt
		]
		\textbf{Input:}\\[0.3em]
		Evaluate this multi-turn function-calling sample against quality criteria.

		\vspace{0.3em}
		\textbf{\# Sample Data}\\
		\texttt{\{sample\_summary\}}
	\end{tcolorbox}

	\vspace{0.3em}

	\begin{tcolorbox}[
			colback=orange!8,
			colframe=orange!60,
			boxrule=1pt,
			arc=3pt,
			left=6pt,
			right=6pt,
			top=4pt,
			bottom=4pt
		]
		\textbf{\# Quality Criteria}
		\begin{enumerate}[leftmargin=1.5em, itemsep=0.2em]
			\item Query-GT Alignment: Does each user query accurately describe what the GT function calls do?
			\item State Consistency: Do parameter values reflect the actual environment config?
			\item Cross-Turn Coherence: Is there logical state progression across turns?
			\item Query Naturalness: Do queries sound like real user requests?
			\item Structural Correctness: Are deliberate ambiguities (e.g., missing parameters) properly reflected?
		\end{enumerate}
	\end{tcolorbox}

	\vspace{0.3em}

	\begin{tcolorbox}[
			colback=red!8,
			colframe=red!60,
			boxrule=1pt,
			arc=3pt,
			left=6pt,
			right=6pt,
			top=4pt,
			bottom=4pt
		]
		\textbf{\# Automatic Rejection Patterns}\\[0.3em]
		If the query contains any of these patterns, REJECT immediately:
		\begin{itemize}[leftmargin=1.5em, itemsep=0.1em]
			\item ``Thought Process'', ``Construct Query'', ``Step 1:'', ``Step 2:''
			\item Function names or parameter names mentioned explicitly
			\item Technical jargon that a real user wouldn't say
		\end{itemize}
	\end{tcolorbox}

	\vspace{0.3em}

	\begin{tcolorbox}[
			colback=toolresponsecolor!8,
			colframe=toolresponsecolor!60,
			boxrule=1pt,
			arc=3pt,
			left=6pt,
			right=6pt,
			top=4pt,
			bottom=4pt
		]
		\textbf{\# Output Format}\\[0.3em]
		First explain your reasoning inside \texttt{<reason></reason>} tags.
		Then output your decision:

		\vspace{0.3em}
		\texttt{<reason> Your detailed analysis... </reason>}\\
		\texttt{<decision>accept</decision>} or \texttt{<decision>reject</decision>}\\
		\texttt{<fail\_reason>Specific reason for rejection (if rejected)</fail\_reason>}
	\end{tcolorbox}

\end{tcolorbox}

\subsection{Refine Classify Agent}
\label{app:refine_classify_prompt}

When the Quality Judge rejects a sample, the Refine Classify Agent determines whether the issue can be fixed by rewriting the user query or if the ground truth itself is unfixable.

\begin{tcolorbox}[
		breakable,
		colback=white,
		colframe=purple!80,
		boxrule=1.5pt,
		arc=4pt,
		left=8pt,
		right=8pt,
		top=6pt,
		bottom=6pt,
		width=0.96\textwidth,
		title={\textbf{Refine Classify Agent System Prompt}},
		fonttitle=\small\bfseries,
		coltitle=white,
		colbacktitle=purple!80
	]
	\small

	\begin{tcolorbox}[
			colback=gray!10,
			colframe=gray!60,
			boxrule=1pt,
			arc=3pt,
			left=6pt,
			right=6pt,
			top=4pt,
			bottom=4pt
		]
		\textbf{System Prompt:}\\[0.3em]
		You are a precise diagnostic assistant. Analyze the root cause of data quality issues.
	\end{tcolorbox}

\end{tcolorbox}

\begin{tcolorbox}[
		breakable,
		colback=white,
		colframe=purple!80,
		boxrule=1.5pt,
		arc=4pt,
		left=8pt,
		right=8pt,
		top=6pt,
		bottom=6pt,
		width=0.96\textwidth,
		title={\textbf{Refine Classify Agent User Prompt}},
		fonttitle=\small\bfseries,
		coltitle=white,
		colbacktitle=purple!80
	]
	\small

	\begin{tcolorbox}[
			colback=gray!10,
			colframe=gray!60,
			boxrule=1pt,
			arc=3pt,
			left=6pt,
			right=6pt,
			top=4pt,
			bottom=4pt
		]
		\textbf{Input:}\\[0.3em]
		A Quality Judge rejected this multi-turn function-calling data sample.

		\vspace{0.3em}
		Rejection reason: \texttt{\{fail\_reason\}}

		\vspace{0.3em}
		Data summary:\\
		\texttt{\{data\_summary\}}
	\end{tcolorbox}

	\vspace{0.3em}

	\begin{tcolorbox}[
			colback=orange!8,
			colframe=orange!60,
			boxrule=1pt,
			arc=3pt,
			left=6pt,
			right=6pt,
			top=4pt,
			bottom=4pt
		]
		\textbf{Task:}\\[0.2em]
		Analyze the rejection reason and determine:
		\begin{itemize}[leftmargin=1.5em, itemsep=0.1em]
			\item Is the problem in the USER QUERY (wrong wording, mentions wrong values, unnatural phrasing, format issues)? $\rightarrow$ These can be fixed by rewriting the query.
			\item Is the problem in the GT function calls (wrong parameters, wrong function, wrong cross-turn state, calling non-existent resources)? $\rightarrow$ These CANNOT be fixed by rewriting the query.
		\end{itemize}
	\end{tcolorbox}

	\vspace{0.3em}

	\begin{tcolorbox}[
			colback=toolresponsecolor!8,
			colframe=toolresponsecolor!60,
			boxrule=1pt,
			arc=3pt,
			left=6pt,
			right=6pt,
			top=4pt,
			bottom=4pt
		]
		\textbf{\# Output Format}\\[0.3em]
		First explain your reasoning inside \texttt{<reason></reason>} tags, then output your answer.

		\vspace{0.3em}
		\texttt{<reason> Your analysis of where the root cause is... </reason>}\\
		\texttt{<answer>query\_fixable</answer>} or \texttt{<answer>gt\_unfixable</answer>}
	\end{tcolorbox}

\end{tcolorbox}

\subsection{Refine Rewrite Agent}
\label{app:refine_rewrite_prompt}

If the Refine Classify Agent determines the issue is query-fixable, the Refine Rewrite Agent performs a targeted rewrite of the problematic user query.

\begin{tcolorbox}[
		breakable,
		colback=white,
		colframe=purple!80,
		boxrule=1.5pt,
		arc=4pt,
		left=8pt,
		right=8pt,
		top=6pt,
		bottom=6pt,
		width=0.96\textwidth,
		title={\textbf{Refine Rewrite Agent System Prompt}},
		fonttitle=\small\bfseries,
		coltitle=white,
		colbacktitle=purple!80
	]
	\small

	\begin{tcolorbox}[
			colback=gray!10,
			colframe=gray!60,
			boxrule=1pt,
			arc=3pt,
			left=6pt,
			right=6pt,
			top=4pt,
			bottom=4pt
		]
		\textbf{System Prompt:}\\[0.3em]
		You are a helpful assistant that rewrites user queries to be natural and accurate.
	\end{tcolorbox}

\end{tcolorbox}

\begin{tcolorbox}[
		breakable,
		colback=white,
		colframe=purple!80,
		boxrule=1.5pt,
		arc=4pt,
		left=8pt,
		right=8pt,
		top=6pt,
		bottom=6pt,
		width=0.96\textwidth,
		title={\textbf{Refine Rewrite Agent User Prompt}},
		fonttitle=\small\bfseries,
		coltitle=white,
		colbacktitle=purple!80
	]
	\small

	\begin{tcolorbox}[
			colback=gray!10,
			colframe=gray!60,
			boxrule=1pt,
			arc=3pt,
			left=6pt,
			right=6pt,
			top=4pt,
			bottom=4pt
		]
		\textbf{Input:}\\[0.3em]
		A quality check found an issue with this user query in a function-calling dataset.

		\vspace{0.3em}
		Issue: \texttt{\{fail\_reason\}}

		\vspace{0.3em}
		Original query: ``\texttt{\{old\_query\}}''\\
		Ground truth function calls for this turn: \texttt{\{gt\_str\}}
	\end{tcolorbox}

	\vspace{0.3em}

	\begin{tcolorbox}[
			colback=orange!8,
			colframe=orange!60,
			boxrule=1pt,
			arc=3pt,
			left=6pt,
			right=6pt,
			top=4pt,
			bottom=4pt
		]
		\textbf{Task:}\\[0.2em]
		Rewrite the user query so that:
		\begin{enumerate}[leftmargin=1.5em, itemsep=0.2em]
			\item It naturally and accurately describes what the GT function calls actually do
			\item It sounds like a real user talking to an AI assistant
			\item It does NOT mention function names, parameter names, or technical details
			\item It does NOT contain special characters that could cause parsing issues
			\item It fixes the specific issue described above
		\end{enumerate}
	\end{tcolorbox}

	\vspace{0.3em}

	\begin{tcolorbox}[
			colback=toolresponsecolor!8,
			colframe=toolresponsecolor!60,
			boxrule=1pt,
			arc=3pt,
			left=6pt,
			right=6pt,
			top=4pt,
			bottom=4pt
		]
		\textbf{\# Output Format}\\[0.3em]
		Output ONLY the rewritten query inside \texttt{<answer>} tags:\\
		\texttt{<answer>your rewritten query here</answer>}
	\end{tcolorbox}

\end{tcolorbox}

\section{Deterministic Execution Pipeline Internals}
\label{app:pipeline_details}

The Stage II execution pipeline instantiates the abstract plan within a Python-based sandbox environment that faithfully replicates the training environment's API semantics.
The pipeline processes each turn sequentially through the following stages:

\begin{enumerate}[nosep, leftmargin=15pt]
	\item \textbf{Function Sampling:} Given the Planner's output specifying which functions to call per turn, the pipeline samples concrete function instances from the available catalog. HIGH-LEVEL functions are automatically decomposed into sequences of BOTTOM-LEVEL API calls.
	\item \textbf{Parameter Generation:} For each selected function, an LLM generates concrete parameter values conditioned on the function schema, the current environment state $\mathcal{C}_t$, and any dependencies from previous turns.
	\item \textbf{VM Execution:} The parameterized function call is executed against the sandbox VM. The VM maintains a complete environment state (file systems, account balances, database entries, etc.) and returns execution results or error messages.
	\item \textbf{Query Generation:} A per-class LLM prompt generates a natural user query that describes the function call's intent without exposing function names or parameters.
	\item \textbf{Query Verification:} A verification LLM checks whether the generated query's semantics align with the GT function calls, rejecting queries that are misaligned or contain data generation artifacts.
\end{enumerate}

If any stage fails, a structured error is recorded with the error type, failing function, turn number, and diagnostic detail. This structured error feeds directly into the Config Patch Agent and the Planner re-invocation described in the main text.

\section{Error Taxonomy for Feedback-Driven Re-synthesis}
\label{app:error_taxonomy}
To provide structured feedback to the Error Critic and Planner Agent, the execution pipeline classifies failures into the following categories. Each error type triggers specific recovery strategies:

\begin{enumerate}[nosep, leftmargin=15pt]
	\item \textbf{\texttt{param\_gen\_failed}:} The LLM failed to generate valid parameters for a function call. The Config Patch Agent may adjust environment state to make parameter generation feasible.
	\item \textbf{\texttt{decompose\_failed}:} A HIGH-LEVEL function could not be decomposed into valid BOTTOM-LEVEL calls. The Planner is instructed to ``Use BOTTOM-LEVEL functions only.''
	\item \textbf{\texttt{func\_sample\_failed}:} No valid function could be sampled after multiple retries (e.g., all candidates require unavailable prerequisites). The Planner is told to ``AVOID functions requiring authentication or specific prior state.''
	\item \textbf{\texttt{vm\_exec\_failed}:} The function call executed but the VM returned an error (e.g., insufficient balance, closed market, missing resource). This is the primary trigger for Config Patching.
	\item \textbf{\texttt{duplicate\_func}:} The same function call appeared multiple times in the same turn. The pipeline rejects the sample.
	\item \textbf{\texttt{query\_gen\_failed}:} The query generation LLM failed to produce a valid \texttt{<query>} tag after multiple retries. The Planner is told to ``Use simpler function combinations.''
	\item \textbf{\texttt{query\_verify\_failed}:} The generated query was rejected by the verification LLM as semantically misaligned with the GT. The Planner is told to ``Use 1 function per turn to simplify.''
	\item \textbf{\texttt{query\_verify\_no\_tag}:} The verification LLM did not return a parseable verdict tag.
	\item \textbf{\texttt{conversation\_construct\_failed}:} The multi-turn conversation assembly failed (e.g., cross-turn dependency resolution error).
	\item \textbf{\texttt{no\_prompts}:} No prompt template exists for the specified class.
	\item \textbf{\texttt{no\_pattern}:} No valid execution pattern could be derived for the given class order.
	\item \textbf{\texttt{pipeline\_exception}:} An uncaught exception during pipeline execution.
\end{enumerate}

Only the first four error types (\texttt{param\_gen\_failed}, \texttt{decompose\_failed}, \texttt{func\_sample\_failed}, \texttt{vm\_exec\_failed}) trigger the Config Patch Agent, as these represent environment-level issues resolvable through state modification. Query-level errors trigger Planner re-invocation with action space constraints but do not invoke config patching.

\section{Feedback Loop Implementation Details}
\label{app:feedback_loop}
This section provides detailed implementation of the feedback-driven re-synthesis mechanism.

\textbf{Dual-Feedback Loop.}
When the execution engine encounters an error (classified per Appendix~\ref{app:error_taxonomy}), recovery proceeds through a dual-feedback loop that simultaneously accumulates two types of corrective signals across $K_{\max}\!=\!3$ pipeline attempts:
\begin{enumerate}[leftmargin=*, nosep]
	\item \textbf{Environment Config Patching:} If the error type is patchable (\texttt{param\_gen\_failed}, \texttt{vm\_exec\_failed}, \texttt{func\_sample\_failed}, or \texttt{decompose\_failed}), the Config Patch Agent (Appendix~\ref{app:config_patch_prompt}) analyzes the initial environment configuration and outputs structured XML patches to update the state. Patches are accumulated across retries via recursive deep-merge, where later patches override earlier ones for the same field but coexist for different fields. A safety mechanism prevents non-dict values from overwriting dict-structured fields (e.g., preventing a string summary from corrupting a file tree).
	\item \textbf{Action Space Pruning:} Simultaneously, the names of all functions involved in failures are extracted and added to a cumulative blocklist. On re-invocation, the Planner Agent receives: (a) the full failure history with structured error descriptions, (b) a list of specifically blocked function names, and (c) error-type-specific guidance (e.g., ``Use BOTTOM-LEVEL functions only'' for decomposition failures, ``AVOID functions requiring authentication'' for sampling failures). The Planner is explicitly instructed to ``Generate a COMPLETELY DIFFERENT plan using different functions.''
\end{enumerate}

Both corrective signals are applied jointly to the next pipeline attempt, progressively narrowing the search space until a valid instantiation is found. If all $K_{\max}$ attempts fail, the seed is discarded.

\section{Quality Judge and Refinement Loop}
\label{app:quality_judge}
This section details the multi-tier validation pipeline and the iterative refinement mechanism.

\textbf{Rule-Based Validation (Gate 1--3).}
Before reaching the LLM Quality Judge, variants must pass three deterministic gates:
\begin{itemize}[nosep, leftmargin=12pt]
	\item \textbf{VM Re-verification:} All GT function calls are re-executed against a fresh VM instance initialized with the variant's \texttt{initial\_config}, ensuring execution correctness.
	\item \textbf{Tool Availability:} Every function name in the GT must exist in the tools provided to the model. For \texttt{miss\_func} variants, this check accounts for tools provided in recovery turns.
	\item \textbf{Parameter Complexity:} List/tuple parameters are limited to $\leq 5$ elements; string parameters to $\leq 200$ characters. This ensures the model can realistically reproduce the GT during training (pass@16 $> 0$).
\end{itemize}

\textbf{LLM Quality Judge (Gate 4).}
The Quality Judge strictly evaluates synthesized trajectories against five semantic criteria:
(1) \textbf{Query--GT Alignment} (ground truth must exactly match the query's intent---no more, no less),
(2) \textbf{State Consistency} (parameter values must reflect the actual environment config, not incorrect values stated by the user),
(3) \textbf{Cross-Turn Coherence} (logical state progression across turns),
(4) \textbf{Query Naturalness} (human-like conversational flow without data generation artifacts), and
(5) \textbf{Structural Correctness} (e.g., ensuring deliberate ambiguity in missing-parameter scenarios).
Automatic failure is triggered by prompt-leakage patterns (e.g., ``Thought Process'', ``Construct Query'', raw JSON tool definitions in non-recovery turns). The judge outputs a \texttt{<reason>} analysis followed by a \texttt{<decision>accept</decision>} or \texttt{<decision>reject</decision>} verdict. Full prompt is provided in Appendix~\ref{app:quality_judge_prompt}.

\textbf{Diagnostic Refinement Loop (max 1 cycle).}
If a trajectory is rejected, a \textbf{Fail Classifier} (Appendix~\ref{app:refine_classify_prompt}) analyzes the rejection reason and outputs one of two verdicts:
\begin{itemize}[nosep, leftmargin=12pt]
	\item \texttt{gt\_unfixable}: The problem lies in the GT function calls (e.g., wrong parameters, wrong function, state dependency violations). The sample is \textbf{immediately dropped}.
	\item \texttt{query\_fixable}: The problem is in the user query (e.g., unnatural wording, misaligned description). The sample is routed to a \textbf{Refine Rewriter} (Appendix~\ref{app:refine_rewrite_prompt}) which performs a one-shot targeted rewrite of the identified turn's query.
\end{itemize}
The rewritten variant is submitted to the Quality Judge for a second evaluation. Trajectories that fail this second check are permanently dropped. No recursive refinement is applied.

\section{Reward Details}
\label{app:reward_details}
This section details the format reward $R_{\text{format}}$ design used in Stage 1 training, which evaluates the structural correctness of tool calls.

\textbf{Format Reward Formulation (Stage 1).}
In Stage 1, the agent is optimized purely for formatting and valid API execution syntax. We assign per-turn penalty codes: $-3$ for XML format errors, $-2$ for tool schema errors (e.g., invalid JSON), and $-1$ for valid syntax but execution failure. Valid executions receive $0$ or $1$. Let $N$ be the total interaction rounds (turns), and $n_{-k}$ be the count of turns receiving code $-k$. The format and tool execution rewards are defined as:
\begin{equation}
	r_{\text{format}} = \max\left(0, \frac{N - n_{-3}}{N}\right), \qquad
	r_{\text{tool}} = \begin{cases}
		\frac{n_{-1}}{n_{-1}+n_{-2}} & \text{if } n_{-1}+n_{-2} > 0 \\
		0                            & \text{otherwise}
	\end{cases}
\end{equation}
An indicator function $\mathbb{1}_{\text{tool}} = \mathbb{I}[n_{-1}+n_{-2} > 0]$ ensures at least one tool call was attempted. The final Stage 1 reward is defined as $R_{\text{final}} = \mathbb{1}_{\text{tool}} \cdot (r_{\text{format}} + r_{\text{tool}}) \in [0, 2]$. In Stages 2 and 3, the Progress Reward ($R_P$) is computed simply as the fraction of successfully resolved turns.

\section{Combination Experiment: RODS + EnvTuning}
\label{app:combination}
To analyze whether boundary-targeted synthesis (\ours) and environmental feedback augmentation (EnvTuning) provide complementary gradient signals, we plot their training progress rewards. As shown in Table~\ref{tab:main_results}, the static baseline without augmentation suffers from rapid reward saturation. While \ours maintains a continuously climbing progress reward by introducing boundary tasks, replacing static tasks alone still lacks granular execution hints. By combining both methods, the agent not only receives high-variance boundary tasks but also environment-aided corrections, leading to improved training stability and sample efficiency.

\section{Training Details}
\label{app:training_details}
This section lists the full training hyperparameters for the RL pipeline.

\begin{table}[htbp]
	\centering
	\caption{\small \textbf{System Hyperparameters.} Configuration details for RL training, data synthesis, and the dynamic replay buffer.}
	\label{tab:hyperparameters}
	\resizebox{0.75\linewidth}{!}{
		\begin{tabular}{llc}
			\toprule
			\textbf{Category} & \textbf{Hyperparameter}                                    & \textbf{Value}     \\
			\midrule
			\multirow{5}{*}{\textbf{RL Training (GRPO)}}
			                  & Actor Learning Rate                                        & $1 \times 10^{-6}$ \\
			                  & KL Loss Coefficient ($\beta$)                              & $0.01$             \\
			                  & Number of Rollouts ($K$)                                   & $16$               \\
			                  & PPO Mini-batch Size                                        & $512$              \\
			                  & Total Training Epochs                                      & $5$ (per stage)    \\
			\midrule
			\multirow{4}{*}{\textbf{Data Synthesis}}
			                  & LLM Backend                                                & Qwen3-32B          \\
			                  & Decoding Temperature                                       & $1.0$              \\
			                  & Top-$p$                                                    & $0.7$              \\
			                  & Max Pipeline / Planner Retries                             & $3$ / $3$          \\
			\midrule
			\multirow{5}{*}{\textbf{Dynamic Buffer}}
			                  & Boundary Lower Bound ($\alpha^-$)                          & $0.20$             \\
			                  & Boundary Upper Bound ($\alpha^+$)                          & $0.85$             \\
			                  & Generated Pool Cap ($P_{\max}$)                            & $400$              \\
			                  & Trial-Period Observation Count                             & $1$                \\
			                  & Trial-Period Eviction Threshold                            & $0.20$             \\
			                  & Retirement Mastered Threshold ($\alpha^+_{\text{retire}}$) & $0.95$             \\
			                  & Retirement Hard Threshold ($\alpha^-_{\text{retire}}$)     & $0.20$             \\
			\bottomrule
		\end{tabular}
	}
\end{table}

\section{Theoretical Extension to PPO}
\label{app:ppo_extension}
Although our empirical results focus on GRPO due to its memory efficiency, the \ours capability probe naturally extends to Proximal Policy Optimization (PPO). In PPO, the advantage function $A_t$ is typically estimated via Generalized Advantage Estimation (GAE), driven by the TD-error $\delta_t = r_t + \gamma V(s_{t+1}) - V(s_t)$.

For binary or bounded task-level rewards (like our Progress Reward $R_P$), the variance of the TD-error across multiple trajectory samples from the same prompt remains tightly coupled to the variance of the final reward. When a task is fully mastered ($R_P \to 1$) or consistently failed ($R_P \to 0$), the Value network $V(s)$ accurately predicts the outcome, leading to near-zero TD-errors ($\delta_t \approx 0$) and vanishing gradients. Conversely, at the capability boundary where outcomes are highly uncertain ($p \approx 0.5$), the Value network's prediction error is maximized, resulting in high-variance advantage estimates.

Therefore, tracking the variance of the PPO advantage estimates $\text{Var}(\hat{A})$ or the raw reward variance inside the PPO rollout buffer serves identical functions to the GRPO variance probe, allowing \ours to dynamically identify and synthesize boundary tasks without architectural changes.

\section{Out-of-Distribution Generalization Results}
\label{app:ood_results}
This section provides the detailed performance metrics for out-of-distribution (OOD) generalization across the BFCL V4, $\tau^2$-bench, and ACEBench Agent benchmarks. The results confirm that boundary-expanded training yields generalizable reasoning capabilities rather than mere in-distribution pattern matching.

\begin{table*}[htbp]
	\centering
	\caption{\small
		\textbf{OOD generalization performance on BFCL V4, $\tau^2$-bench, and ACEBench Agent benchmarks}.
		All results are compared against the \textbf{Llama-3.1-8B-Instruct} base model.
		Models trained with \ours (rows in blue) show improvements on these OOD tasks.
		Scores for xLAM on the Retail and Airline domains are \textcolor{gray}{\textbf{grayed out}} as it was trained on the original $\tau$-bench, making them invalid for OOD evaluation.
		\looseness=-1
	}
	\label{tab:ood_results}
	\resizebox{\textwidth}{!}{
		\begin{tabular}{l ccccccccccc}
			\toprule
			\multirow{2}{*}{\textbf{Model}} & \multicolumn{3}{c}{\textbf{BFCL V4}} & \multicolumn{4}{c}{\textbf{$\tau^2$-bench}} & \multicolumn{3}{c}{\textbf{ACEBench Agent}}                                                                                                                                                                                   \\
			\cmidrule(lr){2-4} \cmidrule(lr){5-8} \cmidrule(lr){9-11}
			                                & \textbf{Avg. (\%)}                   & \textbf{Web Search (\%)}                    & \textbf{Memory (\%)}                        & \textbf{Avg. (\%)}      & \textbf{Retail (\%)}    & \textbf{Airline (\%)}   & \textbf{Telecom (\%)} & \textbf{Avg. (\%)}  & \textbf{Multi-turn (\%)} & \textbf{Multi-step (\%)} \\
			\midrule
			\rowcolor{red!10}
			xLAM-2-8b-fc-r                  & \improvement{9.17}                   & \improvement{5.00}                          & \improvement{13.33}                         & \textcolor{gray}{36.67} & \textcolor{gray}{57.50} & \textcolor{gray}{40.00} & \degradation{12.50}   & \improvement{9.15}  & \improvement{13.30}      & 5.00                     \\
			\rowcolor{red!10}
			BitAgent-8B                     & \improvement{7.41}                   & \improvement{4.50}                          & \degradation{10.32}                         & \degradation{10.00}     & \improvement{7.50}      & \degradation{15.00}     & \degradation{7.50}    & \improvement{8.65}  & \improvement{12.00}      & \improvement{5.30}       \\
			\rowcolor{red!10}
			ToolACE-2-8B                    & \improvement{15.75}                  & \improvement{8.90}                          & \improvement{22.60}                         & \degradation{10.00}     & \improvement{7.50}      & \degradation{15.00}     & \degradation{7.50}    & \improvement{11.00} & \improvement{15.00}      & \improvement{7.00}       \\
			\midrule
			Llama-3.1-8B-Instruct           & 7.05                                 & 2.50                                        & 11.60                                       & 19.46                   & 5.88                    & 30.00                   & 22.50                 & 4.15                & 3.30                     & 5.00                     \\
			\rowcolor{gray!10}
			\, + EnvTuning                  & \improvement{16.53}                  & \improvement{15.00}                         & \improvement{18.06}                         & \improvement{22.17}     & \improvement{6.00}      & \improvement{30.50}     & \improvement{30.00}   & \improvement{8.82}  & \improvement{11.30}      & \improvement{6.33}       \\
			\rowcolor{cyan!8}
			\, + \ours(ours)                & \improvement{18.00}                  & \improvement{15.50}                         & \improvement{20.50}                         & \improvement{25.00}     & \improvement{8.00}      & \improvement{34.00}     & \improvement{33.00}   & \improvement{11.15} & \improvement{14.30}      & \improvement{8.00}       \\
			\bottomrule
		\end{tabular}
	}
\end{table*}

\section{Synthesis LLM Robustness}
\label{app:synthesis_llm_robustness}

A natural question is whether the performance gains of \ours are attributable to the framework design or to the specific generative capacity of the synthesis LLM.
To investigate this, we replace the default synthesis backbone (Qwen3-32B) with GLM-4.5-Air---a generally stronger model---deployed via vLLM on $8\times$A100 GPUs, while keeping all other components---boundary detection, structural isomorphism, dynamic replay buffer, and training pipeline---strictly identical.
The generated pool cap remains $P_{\max}\!=\!400$.

\begin{table}[htbp]
	\centering
	\caption{\small
		\textbf{Synthesis LLM robustness on BFCL V3 multi-turn.}
		Replacing the synthesis backbone with a stronger model (GLM-4.5-Air) yields nearly identical overall performance ($\downarrow$0.75\%), confirming that \ours is model-agnostic: its gains derive from the framework design rather than from the specific synthesis LLM.
	}
	\label{tab:synthesis_llm_robustness}
	\resizebox{0.85\linewidth}{!}{
		\begin{tabular}{l ccccc c}
			\toprule
			\multirow{2}{*}{\textbf{Synthesis LLM}} & \multicolumn{5}{c}{\textbf{BFCL V3 Multi Turn}} & \multirow{2}{*}{$\Delta$\textbf{Avg.}}                                                                               \\
			\cmidrule(lr){2-6}
			                                        & \textbf{Avg.}                                   & \textbf{Base}                          & \textbf{M. Func} & \textbf{M. Param} & \textbf{L. Ctxt} &                   \\
			\midrule
			\rowcolor{cyan!8}
			Qwen3-32B (default)                     & \textbf{56.00}                                  & \textbf{68.00}                         & \textbf{59.00}   & 44.00             & 53.00            & ---               \\
			GLM-4.5-Air                             & 55.25                                           & 65.00                                  & 55.00            & \textbf{46.00}    & \textbf{55.00}   & $\downarrow$ 0.75 \\
			\bottomrule
		\end{tabular}
	}
\end{table}

\paragraph{Analysis.}
The overall performance difference is 0.75\% (56.00\% $\to$ 55.25\%).
GLM-4.5-Air is a stronger model than Qwen3-32B, yet switching to it does not yield further gains.
This suggests limited sensitivity to the synthesis backbone under our setup: schema-guided planning, deterministic VM validation, and multi-tier quality filtering together normalize the output regardless of the underlying LLM's raw capability.
Put differently, \ours's performance is determined by the \emph{framework design} (boundary detection, structural isomorphism, dynamic lifecycle), not by the generative capacity of the synthesis model.

At the sub-split level, the two LLMs show mild distributional differences: GLM-4.5-Air scores higher on \texttt{Missing Parameter} (+2.00\%) and \texttt{Long Context} (+2.00\%), while Qwen3-32B leads on \texttt{Base} (+3.00\%) and \texttt{Missing Functions} (+4.00\%).
These complementary biases suggest that different LLMs produce variants with subtly different structural characteristics, pointing to a potential direction of multi-backbone synthesis ensembles.

\section{Synthesis Computational Cost}
\label{app:synthesis_cost}

We report the computational overhead of the \ours synthesis pipeline to contextualize its cost relative to the RL training loop.

\paragraph{Hardware allocation.}
Training and synthesis run on separate GPU clusters in parallel:
\begin{itemize}[nosep, leftmargin=12pt]
	\item \textbf{RL training:} $8\times$A100 (80GB) GPUs running GRPO with the Qwen3-4B-Instruct policy.
	\item \textbf{Data synthesis:} $8\times$A100 (80GB) GPUs hosting Qwen3-32B via vLLM, with 64 parallel worker threads dispatching synthesis requests.
\end{itemize}
The two systems communicate asynchronously via filesystem-based queues (seed emission from the trainer, variant ingestion at epoch boundaries). The synthesis daemon does not block or slow the training loop.

\paragraph{Wall-clock time.}
To reach the reported performance of 56.00\% (at training step $\sim$600), the system runs for approximately \textbf{56 hours} wall-clock. Since both clusters operate concurrently for the full duration, the total compute is:
\begin{itemize}[nosep, leftmargin=12pt]
	\item Training: $8 \times 56 = 448$ GPU-hours.
	\item Synthesis: $8 \times 56 = 448$ GPU-hours.
	\item \textbf{Total: $\sim$896 GPU-hours} ($16\times$A100 for $\sim$56 hours).
\end{itemize}
The synthesis overhead is thus $1\times$ the training cost in GPU-hours. However, the synthesis cluster runs a single vLLM instance with no gradient computation, so its actual FLOP consumption is lower than the training cluster.

\paragraph{Per-variant synthesis cost.}
On the successful path, generating a single base variant requires approximately \textbf{9--15 LLM calls} depending on the number of turns: 1 Planner call, 1 query generation + 1 query verification per turn (for a 3-turn variant: 6 calls), 1 coherence rewrite, and 1 quality judge evaluation.
For \texttt{miss\_func} and \texttt{miss\_param} variants, 2--5 additional calls are needed for the adversarial transform and its verification.
The 64-worker thread pool processes these calls concurrently across seeds, achieving high throughput despite the per-variant multi-stage pipeline.

\paragraph{Synthesis latency.}
The measured seed-to-injection delay averages \textbf{1 training step} ($\sim$3.7 minutes). Since variant injection occurs at epoch boundaries and the daemon processes seeds continuously in the background, the synthesis pipeline does not introduce any idle time into the training loop. Variants generated during epoch $n$ are staged and injected at the start of epoch $n{+}1$, as described in Section~3.3.

\paragraph{Cost-efficiency perspective.}
While the synthesis overhead doubles the GPU footprint relative to standard GRPO training, the resulting $20\times$ data efficiency gain (matching 17K-sample offline pipelines with an active training pool of $\sim$800 samples) represents a practical trade-off. Generating 17K high-quality multi-turn trajectories offline via similarly complex multi-agent simulation pipelines (such as APIGen-MT or FunReason-MT) typically incurs massive upstream computational costs before training even begins. By explicitly targeting only the high-variance capability boundary, \ours not only reduces the required data volume by an order of magnitude but also provides a favorable end-to-end data-compute trade-off relative to massive offline pipelines.
\section{Benchmark and Evaluation Details}
\label{app:benchmark_details}
We evaluate our agents using several multi-turn tool-use benchmarks to assess both in-distribution learning and out-of-distribution (OOD) generalization.

\paragraph{In-Distribution Evaluation (BFCL V3).}
The Berkeley Function Calling Leaderboard (BFCL) V3~\citep{patil2025bfcl} provides a reliable testbed for multi-turn scenarios. We utilize its 800-sample multi-turn subset, partitioned equally across four categories:
\begin{itemize}[nosep, leftmargin=15pt]
	\item \texttt{Base}: Standard multi-turn tasks with straightforward dependencies.
	\item \texttt{Missing Function}: Tasks requiring the agent to recognize missing capabilities and either gracefully decline or request alternative tools.
	\item \texttt{Missing Parameter}: Tasks lacking necessary arguments, requiring the agent to ask the user for clarification before proceeding.
	\item \texttt{Long-Context}: Scenarios involving extended conversations where context must be maintained across many turns.
\end{itemize}
We adopt the exact 400/400 train/test split established by~\citet{tuneenv} to ensure a fair comparison. Evaluation is performed via the official BFCL abstract syntax tree (AST) matching evaluator.

\paragraph{Out-of-Distribution Evaluation.}
To verify that \ours induces generalizable reasoning rather than mere pattern matching, we evaluate on benchmarks featuring unseen APIs and interaction modalities:
\begin{itemize}[nosep, leftmargin=15pt]
	\item \textbf{BFCL V4:} We test on the \texttt{Web Search} and \texttt{Memory} tracks, representing dynamic information retrieval and long-term state tracking not present in V3.
	\item \textbf{$\tau^2$-bench~\citep{tau2bench}:} A dual-control conversational benchmark set in the \texttt{Retail}, \texttt{Airline}, and \texttt{Telecom} domains, emphasizing highly constrained, real-world business logic.
	\item \textbf{ACEBench~\citep{acebench}:} We utilize the \texttt{Multi-turn} and \texttt{Multi-step} splits of the Agent track to test complex API topologies.
\end{itemize}
All OOD evaluations strictly follow their respective official evaluation protocols and scoring scripts.

\section{Synthesized Data Examples}
\label{app:data_examples}

This section presents seed-to-variant pairs for each data type, demonstrating how \ours preserves structural complexity while generating novel content. For each example, we show the original seed (left/top) and the synthesized variant (right/bottom) side by side.

\subsection{Base Type: VehicleControlAPI (Seed $\to$ Variant)}

\textbf{Seed} (\texttt{multi\_turn\_base\_63}): A 3-turn task involving unit conversion, engine startup with safety checks, and distance estimation.

\begin{tcolorbox}[
		breakable, colback=white, colframe=gray!70, boxrule=1pt, arc=3pt,
		left=6pt, right=6pt, top=4pt, bottom=4pt,
		width=0.96\textwidth,
		title={\textbf{Original Seed}},
		fonttitle=\small\bfseries, coltitle=black, colbacktitle=gray!20
	]
	\small
	\textbf{Turn 1:} ``I require assistance in determining the quantity of gasoline necessary for an extensive journey across California. I currently anticipate needing around 166 liters. How much is that in gallons?''\\[0.2em]
	\textbf{Turn 2:} ``Prior to commencing the drive, kindly initiate the engine, ensuring all doors are securely closed and the parking brake is engaged.''\\[0.2em]
	\textbf{Turn 3:} ``Could you provide me with the approximate distance between San Francisco and Rivermist? Will I be able to get there?''
\end{tcolorbox}

\vspace{0.5em}

\textbf{Synthesized Variant}: Same skill structure (unit conversion $\to$ vehicle status + fuel $\to$ multi-step distance estimation), but different functions and parameters.

\begin{tcolorbox}[
		breakable, colback=white, colframe=toolresponsecolor!80, boxrule=1.5pt, arc=4pt,
		left=8pt, right=8pt, top=6pt, bottom=6pt,
		width=0.96\textwidth,
		title={\textbf{Synthesized Variant}},
		fonttitle=\small\bfseries, coltitle=white, colbacktitle=toolresponsecolor!80
	]
	\small

	\begin{tcolorbox}[colback=gray!10, colframe=gray!60, boxrule=1pt, arc=3pt, left=6pt, right=6pt, top=4pt, bottom=4pt]
		\textbf{Turn 1} --- \emph{Unit Conversion}\\[0.3em]
		\textbf{User:} ``I just checked my car's fuel gauge and it shows 7.8 liters remaining. How many gallons is that?''\\[0.2em]
		\textbf{GT:} \texttt{liter\_to\_gallon(liter=7.8)}
	\end{tcolorbox}

	\vspace{0.3em}

	\begin{tcolorbox}[colback=gray!10, colframe=gray!60, boxrule=1pt, arc=3pt, left=6pt, right=6pt, top=4pt, bottom=4pt]
		\textbf{Turn 2} --- \emph{Vehicle Status + Fuel Operation}\\[0.3em]
		\textbf{User:} ``Please show me my current fuel level and add 4 gallons to the tank for the trip.''\\[0.2em]
		\textbf{GT:} \texttt{displayCarStatus(option='fuel')}, \texttt{fillFuelTank(fuelAmount=4.0)}
	\end{tcolorbox}

	\vspace{0.3em}

	\begin{tcolorbox}[colback=gray!10, colframe=gray!60, boxrule=1pt, arc=3pt, left=6pt, right=6pt, top=4pt, bottom=4pt]
		\textbf{Turn 3} --- \emph{Multi-step Distance + Feasibility}\\[0.3em]
		\textbf{User:} ``I'm driving from Crescent Hollow to Autumnville -- what's the total distance, and can I complete the trip with 630 miles of fuel?''\\[0.2em]
		\textbf{GT:} \texttt{get\_zipcode\_based\_on\_city(city='Crescent Hollow')},\\
		\hspace{2.2em}\texttt{get\_zipcode\_based\_on\_city(city='Autumnville')},\\
		\hspace{2.2em}\texttt{estimate\_distance(cityA='69238', cityB='51479')},\\
		\hspace{2.2em}\texttt{estimate\_drive\_feasibility\_by\_mileage(distance=630.0)}
	\end{tcolorbox}

\end{tcolorbox}

\vspace{0.3em}
\noindent\textbf{Structural preservation:} Both share the pattern \emph{unit conversion (1 call) $\to$ vehicle operation (2 calls) $\to$ distance planning (4 calls with dependency chain)}. The variant uses different city names, conversion direction (liters$\to$gallons vs.\ gallons$\to$liters), and vehicle operations (fuel display vs.\ engine startup), forcing the model to generalize the abstract reasoning pattern.

\subsection{Missing Function Type: GorillaFileSystem (Seed $\to$ Variant)}

\textbf{Seed} (\texttt{multi\_turn\_miss\_func\_38}): A file system task where \texttt{rm} is removed, requiring refusal and recovery.

\begin{tcolorbox}[
		breakable, colback=white, colframe=gray!70, boxrule=1pt, arc=3pt,
		left=6pt, right=6pt, top=4pt, bottom=4pt,
		width=0.96\textwidth,
		title={\textbf{Original Seed}},
		fonttitle=\small\bfseries, coltitle=black, colbacktitle=gray!20
	]
	\small
	\textbf{Turn 1:} ``I've misplaced a vital document. Assist in locating a file named `findings\_report' within `SuperResearch'. Could you remove it and the directory.''\\[0.2em]
	\textbf{Turn 2:} \emph{[Function \texttt{rm} removed; agent must refuse]}\\[0.2em]
	\textbf{Turn 3:} ``What's left in the current directory including the hidden files?''
\end{tcolorbox}

\vspace{0.5em}

\textbf{Synthesized Variant}: Different file operations, \texttt{mkdir} removed instead of \texttt{rm}, recovery in Turn 4.

\begin{tcolorbox}[
		breakable, colback=white, colframe=tracecolor!80, boxrule=1.5pt, arc=4pt,
		left=8pt, right=8pt, top=6pt, bottom=6pt,
		width=0.96\textwidth,
		title={\textbf{Synthesized Variant}},
		fonttitle=\small\bfseries, coltitle=white, colbacktitle=tracecolor!80
	]
	\small

	\begin{tcolorbox}[colback=gray!10, colframe=gray!60, boxrule=1pt, arc=3pt, left=6pt, right=6pt, top=4pt, bottom=4pt]
		\textbf{Turn 1:} ``I need to rename my JSON file `wqmmw.json' to `data.csv' for compatibility.''\\[0.2em]
		\textbf{GT:} \texttt{mv(source='wqmmw.json', destination='data.csv')}
	\end{tcolorbox}

	\vspace{0.3em}

	\begin{tcolorbox}[colback=gray!10, colframe=gray!60, boxrule=1pt, arc=3pt, left=6pt, right=6pt, top=4pt, bottom=4pt]
		\textbf{Turn 2:} ``Please check how many lines are in `data.csv' so I can validate the data.''\\[0.2em]
		\textbf{GT:} \texttt{wc(file\_name='data.csv', mode='l')}
	\end{tcolorbox}

	\vspace{0.3em}

	\begin{tcolorbox}[colback=red!8, colframe=red!60, boxrule=1pt, arc=3pt, left=6pt, right=6pt, top=4pt, bottom=4pt]
		\textbf{Turn 3} --- \emph{\texttt{mkdir} removed from tool list}\\[0.3em]
		\textbf{User:} ``I'd like to create a new folder called `docs' to organize these files.''\\[0.2em]
		\textbf{GT:} \texttt{[]} \hfill (\emph{agent must refuse})
	\end{tcolorbox}

	\vspace{0.3em}

	\begin{tcolorbox}[colback=toolresponsecolor!8, colframe=toolresponsecolor!60, boxrule=1pt, arc=3pt, left=6pt, right=6pt, top=4pt, bottom=4pt]
		\textbf{Turn 4} --- \emph{Function restored; agent recovers}\\[0.3em]
		\textbf{User:} \texttt{[\{``name'': ``mkdir'', ...\}]} ``Here's a tool that might help.''\\[0.2em]
		\textbf{GT:} \texttt{mkdir(dir\_name='docs')}
	\end{tcolorbox}

\end{tcolorbox}

\subsection{Missing Parameter Type: TradingBot + MathAPI (Seed $\to$ Variant)}

\textbf{Seed} (\texttt{multi\_turn\_miss\_param\_144}): A cross-class task where the user provides a vague computation request, requiring parameter clarification.

\begin{tcolorbox}[
		breakable, colback=white, colframe=gray!70, boxrule=1pt, arc=3pt,
		left=6pt, right=6pt, top=4pt, bottom=4pt,
		width=0.96\textwidth,
		title={\textbf{Original Seed}},
		fonttitle=\small\bfseries, coltitle=black, colbacktitle=gray!20
	]
	\small
	\textbf{Turn 1:} ``After determining the current market status, retrieve the stock information for symbol `AAPL'.''\\[0.2em]
	\textbf{Turn 2:} ``Using the current details of a stock, calculate the average of price, trading volume, MA5, and MA20.'' \emph{[Parameters vague -- which stock?]}\\[0.2em]
	\textbf{Turn 3:} ``The stock should be AAPL.'' \emph{[User provides clarification]}
\end{tcolorbox}

\vspace{0.5em}

\textbf{Synthesized Variant}: Different stock, different vague reference (``those two percentage changes''), same clarification pattern.

\begin{tcolorbox}[
		breakable, colback=white, colframe=groundtruthcolor!80, boxrule=1.5pt, arc=4pt,
		left=8pt, right=8pt, top=6pt, bottom=6pt,
		width=0.96\textwidth,
		title={\textbf{Synthesized Variant}},
		fonttitle=\small\bfseries, coltitle=white, colbacktitle=groundtruthcolor!80
	]
	\small

	\begin{tcolorbox}[colback=gray!10, colframe=gray!60, boxrule=1pt, arc=3pt, left=6pt, right=6pt, top=4pt, bottom=4pt]
		\textbf{Turn 1} --- \emph{Stock lookup (TradingBot)}\\[0.3em]
		\textbf{User:} ``I'm looking at Synex Solutions' stock -- can you get their ticker symbol and the latest details?''\\[0.2em]
		\textbf{GT:} \texttt{get\_symbol\_by\_name(name='Synex Solutions')}, \texttt{get\_stock\_info(symbol='SYNX')}
	\end{tcolorbox}

	\vspace{0.3em}

	\begin{tcolorbox}[colback=red!8, colframe=red!60, boxrule=1pt, arc=3pt, left=6pt, right=6pt, top=4pt, bottom=4pt]
		\textbf{Turn 2} --- \emph{Vague query; concrete values omitted}\\[0.3em]
		\textbf{User:} ``I need the average of those two percentage changes we just saw.''\\[0.2em]
		\textbf{GT:} \texttt{[]} \hfill (\emph{agent must ask for clarification})
	\end{tcolorbox}

	\vspace{0.3em}

	\begin{tcolorbox}[colback=toolresponsecolor!8, colframe=toolresponsecolor!60, boxrule=1pt, arc=3pt, left=6pt, right=6pt, top=4pt, bottom=4pt]
		\textbf{Turn 3} --- \emph{User provides missing numerical values}\\[0.3em]
		\textbf{User:} ``They are $-3.4$ and $-1.0$.''\\[0.2em]
		\textbf{GT:} \texttt{mean(numbers=[-3.4, -1.0])}
	\end{tcolorbox}

\end{tcolorbox}

\vspace{0.3em}
\noindent\textbf{Key observation:} In both seed and variant, the ambiguity arises from a vague back-reference to prior tool output. The variant changes the specific stock, the nature of the computation (average of percentage changes vs.\ average of multiple metrics), and the exact missing values, while preserving the core skill: recognizing under-specified parameters and requesting clarification before executing.

\section{Full Benchmark Results}
\label{app:full_results}
\begin{table*}[t]
	\small
	\centering
	\resizebox{\textwidth}{!}{
		\begin{tabular}{@{}lcccccc@{}}
			\toprule
			\textbf{Model}
			                            & \textbf{Size}
			                            & \textbf{Overall}
			                            & \textit{Base}
			                            & \begin{tabular}[c]{@{}c@{}}\textit{Miss}\\ \textit{Func}\end{tabular}
			                            & \begin{tabular}[c]{@{}c@{}}\textit{Miss}\\ \textit{Param}\end{tabular}
			                            & \begin{tabular}[c]{@{}c@{}}\textit{Long}\\ \textit{Context}\end{tabular}                                                                                                                                                                        \\
			\midrule
			\rowcolor{gray!15}
			\multicolumn{7}{l}{\emph{\textbf{Closed-source Model}}}                                                                                                                                                                                                                       \\[1pt]
			Claude-Sonnet-4-5-20250929  & -                                                                        & 61.38                          & 69.00                           & 65.00                          & 52.50                           & 59.00                          \\
			Claude-Haiku-4-5-20251001   & -                                                                        & 53.63                          & 63.50                           & 42.50                          & 52.50                           & 56.00                          \\
			Gemini-3-Pro-Preview        & -                                                                        & 60.75                          & 64.50                           & 60.00                          & 54.50                           & 64.00                          \\
			Gemini-2.5-Pro-Preview      & -                                                                        & 28.75                          & 32.00                           & 29.00                          & 22.00                           & 32.00                          \\
			Gemini-2.5-Flash            & -                                                                        & 16.75                          & 14.50                           & 16.50                          & 17.50                           & 18.50                          \\
			Grok-4-1-fast-reasoning     & -                                                                        & 58.88                          & 70.50                           & 59.50                          & 43.00                           & 62.50                          \\
			Grok-4-1-fast-non-reasoning & -                                                                        & 46.75                          & 58.00                           & 39.50                          & 37.50                           & 52.00                          \\
			GPT-5.2-2025-12-11          & -                                                                        & 28.13                          & 36.50                           & 18.00                          & 27.50                           & 30.50                          \\
			GPT-4o-2024-11-20           & -                                                                        & 42.50                          & 55.50                           & 34.50                          & 29.00                           & 51.00                          \\
			\midrule
			\rowcolor{gray!15}
			\multicolumn{7}{l}{\emph{\textbf{Open-source Model}}}                                                                                                                                                                                                                         \\[1pt]
			Kimi-K2-Instruct            & 1043B                                                                    & 50.63                          & 62.00                           & 41.00                          & 44.50                           & 55.00                          \\
			DeepSeek-V3.2-Exp           & 671B                                                                     & 44.88                          & 55.00                           & 49.00                          & 27.00                           & 48.50                          \\
			Llama-4-Maverick            & 400B                                                                     & 20.25                          & 27.00                           & 22.00                          & 14.00                           & 18.00                          \\
			Qwen3-235B-A22B-Instruct    & 235B                                                                     & 44.63                          & 54.00                           & 42.50                          & 31.50                           & 50.50                          \\
			Qwen3-32B                   & 32B                                                                      & 47.88                          & 56.00                           & 52.50                          & 40.00                           & 43.00                          \\
			Qwen3-30B-A3B-Thinking      & 30B                                                                      & 30.00                          & 43.50                           & 10.50                          & 25.00                           & 41.00                          \\
			ToolACE-2-8B                & 8B                                                                       & 38.38                          & 49.00                           & 28.00                          & 30.50                           & 46.00                          \\
			ToolACE-MT                  & 8B                                                                       & 40.25                          & 57.50                           & 31.50                          & 34.00                           & 38.00                          \\
			Nanbeige4-3B-Thinking-2511  & 3B                                                                       & 51.12                          & 58.50                           & 54.00                          & 45.00                           & 47.00                          \\
			xLAM-2-3b-fc-r              & 3B                                                                       & 58.38                          & 71.50                           & 59.00                          & 57.50                           & 45.50                          \\
			\midrule
			Qwen2.5-7B-Instruct         & 7B                                                                       & $7.00$                         & $9.33$                          & $9.33$                         & $6.33$                          & $3.00$                         \\
			\rowcolor{gray!10}
			\, + Static dataset         & -                                                                        & $36.92$ \improvement{(+29.92)} & $50.33$ \improvement{(+41.00)}  & $40.33$ \improvement{(+31.00)} & $29.33$ \improvement{(+23.00)}  & $27.67$ \improvement{(+24.67)} \\
			\rowcolor{blue!10}
			\, + EnvTuning              & -                                                                        & $37.75$ \improvement{(+30.75)} & $51.50$ \improvement{(+42.17)}  & $41.00$ \improvement{(+31.67)} & $30.50$ \improvement{(+24.17)}  & $28.00$ \improvement{(+25.00)} \\
			\, + \ours(ours)            & -                                                                        & $40.25$ \improvement{(+33.25)} & $54.00$ \improvement{(+44.67)}  & $43.50$ \improvement{(+34.17)} & $33.50$ \improvement{(+27.17)}  & $30.00$ \improvement{(+27.00)} \\
			\hdashline
			Llama-3.1-8B-Instruct       & 8B                                                                       & $5.48$                         & $6.15$                          & $6.80$                         & $3.20$                          & $5.75$                         \\
			\rowcolor{gray!10}
			\, + Static dataset         & -                                                                        & $28.25$ \improvement{(+22.77)} & $28.20$ \improvement{(+22.05)}  & $25.85$ \improvement{(+19.05)} & $22.15$ \improvement{(+18.95)}  & $36.80$ \improvement{(+31.05)} \\
			\rowcolor{blue!10}
			\, + EnvTuning              & -                                                                        & $28.38$ \improvement{(+22.90)} & $28.00$  \improvement{(+21.85)} & $25.50$ \improvement{(+18.70)} & $23.00$  \improvement{(+19.80)} & $37.00$ \improvement{(+31.25)} \\
			\rowcolor{cyan!8}
			\, + \ours(ours)            & -                                                                        & $30.88$ \improvement{(+25.40)} & $32.00$ \improvement{(+25.85)}  & $28.00$ \improvement{(+21.20)} & $24.50$ \improvement{(+21.30)}  & $39.00$ \improvement{(+33.25)} \\
			\hdashline
			Qwen3-4B-Instruct           & 4B                                                                       & $22.13$                        & $26.50$                         & $21.00$                        & $15.50$                         & $25.50$                        \\
			\rowcolor{gray!10}
			\, + Static dataset         & -                                                                        & $50.00$ \improvement{(+27.87)} & $62.00$ \improvement{(+35.50)}  & $51.00$ \improvement{(+30.00)} & $35.00$ \improvement{(+19.50)}  & $52.00$ \improvement{(+26.50)} \\
			\rowcolor{blue!10}
			\, + EnvTuning              & -                                                                        & $50.50$ \improvement{(+28.37)} & $64.00$ \improvement{(+37.50)}  & $52.00$ \improvement{(+31.00)} & $35.00$ \improvement{(+19.50)}  & $51.00$ \improvement{(+25.50)} \\
			\rowcolor{cyan!8}
			\, + \ours(ours)            & -                                                                        & $56.00$ \improvement{(+33.87)} & $68.00$ \improvement{(+41.50)}  & $59.00$ \improvement{(+38.00)} & $44.00$ \improvement{(+28.50)}  & $53.00$ \improvement{(+27.50)} \\
			\hdashline
			\bottomrule
		\end{tabular}
	}
	\vspace{-0.5em}
	\caption{\label{tab:full_results}
		\textbf{Full in-distribution performance on BFCL V3 multi-turn} (Tier 1: controlled RL comparisons).
		All RL methods share the same 400 training samples and GRPO setup; only the data/environment strategy differs.
		\improvement{Red text} indicates improvement over the base model.
		The best result within each model group is \textbf{bolded}.
	}
	\vspace{-1em}
\end{table*}

\section{Ablation Configuration Details}
\label{app:ablation_details}
This section provides detailed descriptions of each ablation condition in Table~\ref{tab:ablation}.

\paragraph{(a) Boundary detection ablations.}
\begin{itemize}[leftmargin=12pt]
	\item \textbf{w/ random seed selection:} Instead of selecting seeds from the boundary region ($\mathcal{D}_{\text{boundary}}$), we sample seeds uniformly at random from the entire training pool regardless of their reward. This isolates the effect of boundary-targeted capability tracking.
	\item \textbf{w/ binary acc instead of progress reward:} We replace the continuous Progress Reward $R_P$ with binary task accuracy (1 if all turns correct, 0 otherwise) for boundary detection. This tests whether the fine-grained partial credit of $R_P$ is necessary for accurate boundary identification.
\end{itemize}

\paragraph{(b) Synthesis pipeline ablations.}
\begin{itemize}[leftmargin=12pt]
	\item \textbf{w/o coherence rewrite:} We skip Stage III (holistic semantic grounding). Per-turn queries are generated independently without the Rewrite Agent's narrative-driven single-pass rendering.
	\item \textbf{w/o narrative planning:} The Planner Agent generates a function sequence without an underlying narrative ($\mathcal{N}$). This removes the cross-turn thematic coherence that anchors all turns to a unified goal.
	\item \textbf{w/o feedback loop (blind retry):} We remove the Error Critic and Config Patch Agent. When execution fails, the pipeline simply retries with a fresh random plan rather than accumulating corrective signals.
\end{itemize}

\paragraph{(c) Lifecycle management ablations.}
\begin{itemize}[leftmargin=12pt]
	\item \textbf{w/o retirement mechanism:} All three retirement layers (L1--L3) are disabled. The pool only grows and is never pruned.
	\item \textbf{w/ static pool (no dynamic refresh):} We generate variants once at the beginning of Stage 3 and freeze the pool thereafter. No new variants are synthesized as training progresses.
\end{itemize}

\end{document}